%% file: main.tex
\pdfoutput=1

\documentclass[11pt]{article}

\usepackage{ACL2023}

\usepackage{times}
\usepackage{latexsym}

\usepackage[T1]{fontenc}

\usepackage[utf8]{inputenc}

\usepackage{microtype}

\usepackage{inconsolata}

\usepackage{booktabs}

\usepackage{framed}
\usepackage{graphicx}
\usepackage{enumitem}
\usepackage{everypage}

\usepackage{amsmath}
\usepackage{amssymb}
\usepackage[normalem]{ulem}

\usepackage{xspace}

\usepackage{tcolorbox}
\usepackage{listings}

\usepackage{csquotes}

\usepackage{setspace}

\usepackage{xcolor}
\definecolor{ColorEditAdd}{RGB}{0, 160, 80}
\definecolor{ColorEditDelete}{RGB}{231, 92, 88}
\definecolor{agentyellow}{RGB}{160, 127, 0}
\definecolor{agentgreen}{RGB}{50, 127, 50}
\definecolor{expertgreen}{RGB}{0, 160, 80}

\newcommand{\reviewtext}[1]{{\textit{"\textcolor{teal}{#1}"}}}

\newcommand{\note}[2]{{\color{#1}{#2}}}
\renewcommand{\note}[2]{}

\newcommand{\dataset}{ARIES\xspace}
\newcommand{\margfam}{MARG\xspace}
\newcommand{\margmain}{\mbox{MARG-S}\xspace}

\newcommand{\agentmsg}[3]{\noindent\textbf{#1:} \textcolor{#2}{\small #3}\\}

\newenvironment{reviewgen_prompt}[1]{
    \begin{tcolorbox}[title={\textbf{#1}},colback=white,colframe=black,arc=0mm,boxrule=0.5pt]
    \begin{singlespace}
}{
    \end{singlespace}
    \end{tcolorbox}
}

\author{
Mike D'Arcy$^1$, Tom Hope$^{2,3}$, Larry Birnbaum$^1$, Doug Downey$^{1,3}$ \\
$^1$Northwestern University, $^2$The Hebrew University of Jerusalem,\\
$^3$Allen Institute for AI,  \\
\texttt{m.m.darcy@u.northwestern.edu, tomh@allenai.org,}\\
\texttt{l-birnbaum@northwestern.edu, dougd@allenai.org}
}

\title{{\margfam}: Multi-Agent Review Generation for Scientific Papers}

\begin{document}

\maketitle

\begin{abstract}

We study the ability of LLMs to generate feedback for scientific papers and develop {\margfam},\footnote{\url{https://github.com/allenai/marg-reviewer}} a feedback generation approach using multiple LLM instances that engage in internal discussion.
By distributing paper text across agents, {\margfam} can consume the full text of papers beyond the input length limitations of the base LLM, and by specializing agents and incorporating sub-tasks tailored to different comment types (experiments, clarity, impact) it improves the helpfulness and specificity of feedback.
In a user study, baseline methods using GPT-4 were rated as producing generic or very generic comments more than half the time, and only 1.7 comments per paper were rated as good overall in the best baseline.  Our system substantially improves the ability of GPT-4 to generate specific and helpful feedback, reducing the rate of generic comments from 60\% to 29\% and generating 3.7 good comments per paper (a 2.2x improvement).
\end{abstract}

\input{review_generation_chapter.tex}

\clearpage\phantomsection 

\begingroup
\setlength{\emergencystretch}{1em}
\def\UrlBreaks{\do\/\do-\do\:}
\bibliography{aries,reviewgen}
\bibliographystyle{acl_natbib}
\endgroup

\appendix

\input{review_generation_appendix.tex}

\end{document}

%% file: review_generation_chapter.tex
\section{Introduction}

In recent years, the capabilities of large language models (LLMs) have advanced dramatically, resulting in modern models such as GPT-4 that can perform comparably to humans in some tasks \citep{openai2023gpt4}.  
These advancements provide hope that LLMs may be able to assist human researchers with their writing \citep{fok2023writing,shen2023summarization,mahlow_writing_2023,gmeiner_dimensions_2023}; however, most modern LLMs can only consume limited amounts of text and are primarily trained on non-technical text such as news articles and websites. The ability of models to comprehend and produce long, highly technical text---such as that of scientific papers---remains under-explored.

In this paper, we study the task of automatically generating actionable peer-review feedback for a scientific paper.  This task comprises several reasoning challenges: a reviewer must understand the intent and significance of a work, the technical details of the methodology, and the nuances of how an experiment or proof can be claimed to support a particular conclusion.  They must then identify the ways in a which a paper does or does not fall short and articulate suggestions for improvement.

Modern large language models (LLMs) face a technical challenge in addition to the reasoning challenges involved in generating reviews: namely, they are limited in the total number of tokens they can effectively reason over at once.  As scientific papers can be quite long (thousands or tens of thousands of tokens, in our case), there are many cases in which it is not even possible to provide the whole paper in the model's input.  Even for models that technically support large inputs, they often cannot use the full capacity effectively in practice \citep{qin_nlp_2023,liu_lost_2023}.

We propose multi-agent review generation (\textbf{\margfam}), a method for generating peer-review feedback by prompting an LLM (GPT-4).  We find that by using multiple instances of GPT (hereinafter referred to as "agents"), giving each a portion of the paper, and allowing the agents to communicate with each other, it is possible to generate feedback across the whole paper.  We additionally find that by including aspect-specific "expert"  GPT agents to separately assist with generating comments on experiments, clarity, and impact, the method can perform significantly better than when having a lone agent attempt to generate all types of feedback at once; we refer to this specialized variant of our method as \textbf{\margmain}.

In a user study, {\margmain} generated 3.7 "good" comments per paper (rated by users), whereas a simple baseline of having a single agent generate all comments generated only 1.7 good comments, and a recently proposed method \citep{liang_can_2023} produced only 0.3.   In addition, we found that while users perceived the majority of the comments generated by the baselines as being generic, the vast majority (71\%) of {\margmain}'s comments were rated as specific.  Finally, we analyze the weaknesses of {\margmain}, including high cost and internal communication errors (e.g., failing to include key information in some messages), and suggest directions for future work.

In summary, our contributions are as follows:
\begin{itemize}
    \item We propose a novel method ({\margfam}) that can generate high-quality peer-review feedback even for papers longer than the context size of the base model.
    
    \item We evaluate the quality of our generated feedback against two baselines, using both automatic metrics and a user study.  We find that our method outperforms the strongest baseline by 6.1 recall points in the automated evaluation and generates 2.2x as many helpful comments per review in the user study.
    \item We conduct a thorough analysis of the generated feedback, finding that our proposed method preserves accuracy while generating much more specific comments.
\end{itemize}

\section{Related work}

\subsection{Review generation}

There has been a variety of work that aims to score or improve papers in specific aspects, such as checking statistical tests \citep{nuijten_statcheck_2020}, plagiarism detection \citep{kalnins_screening_2015}, citation recommendation \citep{ali_deep_2020}, and review score prediction \citep{basuki_quality_2022, bharti_peerrec_2023}, among others \citep{kousha_artificial_2023}.  While these are useful tools, they are limited in scope compared to the breadth of feedback authors receive from a real review; our work aims to produce free-form textual review comments across a variety of aspects.

Past work on automatic review generation primarily does so using (relatively) small models that cannot consume the full text of a paper \citep{yuan_kid_review_2022} or use template-filling instead of generating nuanced free-form comments \citep{wang_reviewrobot_2020}.  More recent work has explored using GPT-4 \citep{openai2023gpt4} to verify author checklists \citep{liu_reviewergpt_2023}, but this limits the variety in generated comment types.

Impressona \citep{benharrak_writer_defined_2023} is an editor that allows writers to create AI personas (via GPT-4) to write comments on their work; this is valuable for personalization of feedback, but doesn't focus on finding good techniques and prompts for scientific review generation, and doesn't explore LM-LM interactions; as we show, a simple prompt (akin to what a user might try initially) does poorly on our task compared to our method.

Contemporaneously with our work, \citet{liang_can_2023} conducted a large user study of review generation using GPT-4, finding that GPT-4 could generate helpful review comments.  However, that work simply truncated long papers and did not attempt to address the input size limitations of GPT-4.  In addition, they used a single prompt rather than attempting to construct specialized prompts and "experts" for different comment types, as we do.  We compare our proposed method to that of \citet{liang_can_2023} and find that while their approach is more efficient, ours produces more helpful comments.

\subsection{Multi-agent modeling}

In games and robotics tasks, where there are often distinct roles being performed or multiple physical agents operating in the same environment, various problem-solving algorithms and reinforcement learning techniques have been studied to enable cooperation between agents \citep{zhang_multi_agent_2021,oroojlooy_review_2022}.  Not all of these use communication for cooperation, and those that do typically exchange symbols or vectors rather than natural-language messages.

Recent work has explored multi-persona interaction with prompted LLMs to simulate artificial societies \citep{li_camel_2023,park_generative_2023} and to improve reasoning abilities \citep{du_improving_2023,wang_unleashing_2023}, but this work does not explore the use of multi-agent modeling to scale input size limits and does not investigate their potential for highly technical tasks like scientific review generation.

Contemporaneously with our work, \citet{hong_metagpt_2023} and \citet{wu_autogen_2023} have proposed general frameworks for multi-agent modeling with large language models such as GPT.  \citet{wang_unleashing_2023} has also proposed multi-persona collaboration as a way to improve LLM creativity, although they do not investigate the ability of multi-agent modeling to scale input size limits.  However, none of these works explore review generation applications.

\subsection{LLM context management}

One advantage of multi-agent modeling is to circumvent the input-size limitations of LLMs, which are often prohibitive for long documents.  A variety of other techniques have been investigated in prior work.

Several works have proposed methods for modifying LLM architectures in order to increase the effective input size by using alternative attention formulations \citep{wang_linformer_2020,kitaev_reformer_2020,beltagy_longformer_2020,ivgi_efficient_2022} or incorporating memory retrieval \citep{wu_memorizing_2022}.  However, architecture changes often cannot be applied without retraining models from scratch, and powerful LLMs such as GPT are sometimes available only through a fixed API that does not allow low-level model modifications.  This motivates us to explore techniques that can be applied without changing the underlying model.

Recently, there has been work exploring context management in LLMs by having models summarize a large input one chunk at a time and then operate on the concatenation of the summaries \citep{wu_recursively_2021}, recursively summarize their input/output history to compress it \citep{wang_recursively_2023}, or incorporate retrieval \citep{xu_retrieval_2023,bai_longbench_2023}.  These strategies are effective when only part of the input is needed or when it is clear in advance what details will be important; however, in our review generation task, a paper's shortcomings may involve nuanced details that would be lost with extraction or summarization techniques, so we divide the input among multiple agents that collectively retain the full text throughout the task.

\section{Task definition}

We formulate our task as follows: given a scientific paper, generate a list of {\em actionable} feedback comments that could help authors to improve the paper.  Actionable feedback is defined the same way as in \citet{darcy_aries_2023}; that is, we focus on suggestions and criticism (including implied suggestions---e.g., a question might imply a need for clarification in the paper) rather than positive remarks (e.g., \reviewtext{The paper is sound and of certain interest}).  In addition, we focus on substantive comments rather than simple grammatical or stylistic errors.

In both our multi-agent approach and our simple baseline, a paper is split into chunks of text so that each chunk can fit into the model's input.  The splits are made on paragraph boundaries to avoid breaking sentences, and when presenting the text to the model we annotate each paragraph with its position in the paper (paragraph 1, 2, 3, etc) and the name of the section it appears in.

We note that the input format we use does not include figures or tables (as GPT-4 is a pure language model,\footnote{OpenAI has recently released a vision-enabled version of GPT-4, but this was not available to us at the time of this work.} it cannot consume this information), and many equations are garbled or incomplete due to parsing limitations.  Nonetheless, we expect that many comments can be identified from the text alone, as the main conclusions from tables and figures are often stated in text.

\section{Multi-agent review generation}
\label{sec:reviewgen_multi_agent}

\begin{figure*}[ht]
    \centering
    \begin{minipage}{0.95\textwidth}
        \centering
        \includegraphics[width=\textwidth]{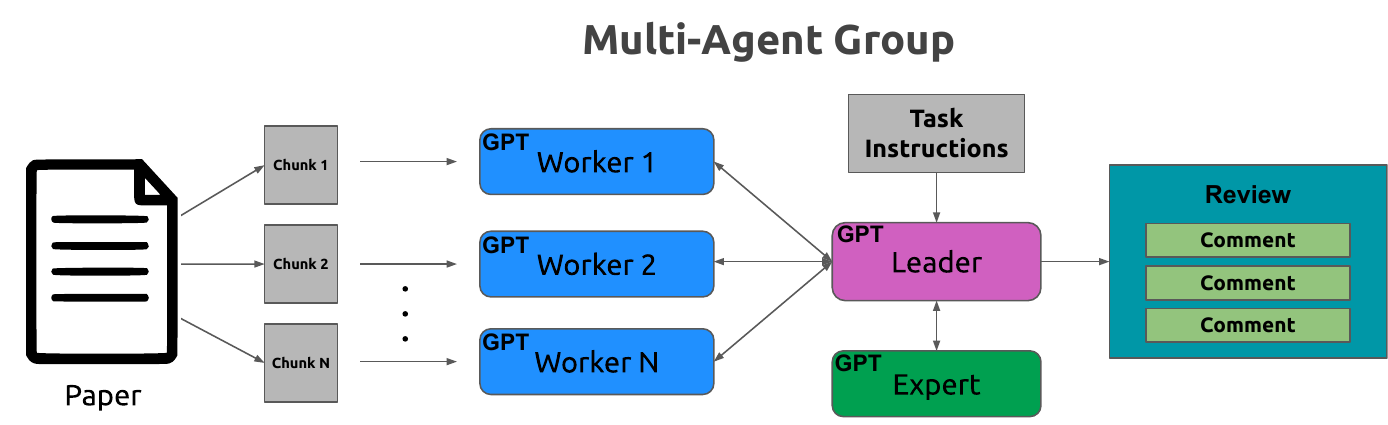}
    \end{minipage}
    \caption{Overview of our multi-agent architecture.}
    \label{fig:marg_comm_overview}
\end{figure*}

\begin{figure*}[ht]
    \centering
    \begin{minipage}{0.95\textwidth}
        \centering
        \includegraphics[width=\textwidth]{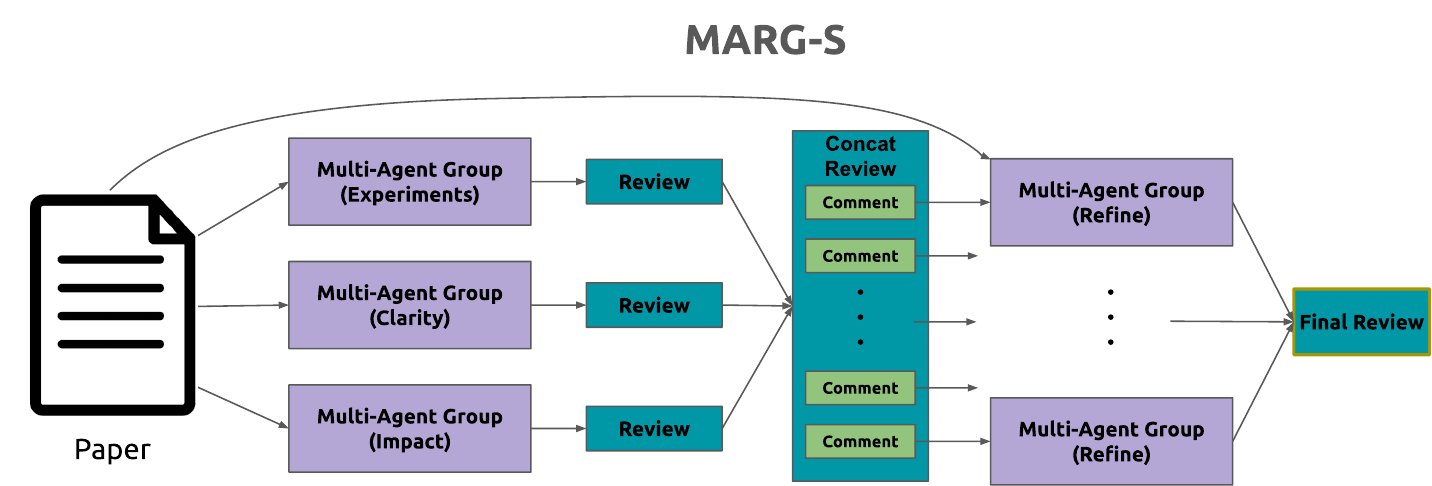}
    \end{minipage}
    \caption{Overview of {\margmain}, which consists of several specialized multi-agent groups.  The comments from each group are concatenated to produce the overall review, and each comment is refined (and potentially pruned) by an additional multi-agent group to produce the final review.}
    \label{fig:marg_s_overview}
\end{figure*}

In this section, we describe our proposed multi-agent method for generating peer-review feedback, which we call {\margmain} (\textbf{M}ulti-\textbf{A}gent \textbf{R}eview \textbf{G}eneration with \textbf{S}pecialized Agents).  At a high level, our multi-agent architecture is formulated as follows: We define an {\em agent} as one instance of a chat-based LLM (ChatGPT, in our case); each agent has its own chat history and prompt(s).  We initialize a set of agents, including three distinct types: (1) a \textbf{\textcolor{magenta}{leader}} agent, which is in charge of coordinating the task and the communication among agents, (2) one or more \textbf{\textcolor{blue}{worker}} agents, which each receive a chunk of the task data (the paper), and (3) zero or more \textbf{\textcolor{expertgreen}{expert}} agents, which are prompted to specialize in some sub-task that serves to assist the leader agent in performing the task effectively.  The leader agent is given a protocol with which it can broadcast a message to all other agents and receive responses.  Finally, the leader agent is given the task instructions, and must send messages to other agents in order to obtain information and delegate sub-tasks in order to produce the final output.

Prompts for all of our methods can be found in \autoref{appendix:reviewgen_prompts}.  An overview of our multi-agent architecture is shown in \autoref{fig:marg_comm_overview}.

\subsection{Agents}

Chat-based LLMs, such as ChatGPT, take an input that consists of a list of messages.  Each message consists of the message content and the "role" that the message is from, which in ChatGPT's case can be either the "system", the "user", or the "assistant" (i.e., generated by the LLM).  Typically, an input to the model starts with a "system" message that describes general instructions that it must follow (e.g., "always give concise and helpful answers"), then the "user" writes a message ("summarize the following passage: ..."), and the generated response is treated as an "assistant" message.  The message history serves as a form of context management; with it, agents can use information from previous interactions in the conversation when formulating future responses.

We use the "system" message at the start of a message history to give unique instructions to each agent type.  For example, the "leader" agent is told that it is the leader, that it must coordinate other agents to complete the user's requests, and that it can communicate by using a special "\texttt{SEND MESSAGE}" command to broadcast messages to other agents.  It is also given some guidelines to improve its reasoning; for example, it is instructed to create a high-level plan from its task instructions before it begins communicating and performing sub-tasks.  The "worker" agents are told that they must obey instructions from the leader agent, and "experts" are given special instructions depending on the sub-task they need to perform.

Despite their name, expert agents do not actually have more information or expertise than any other agent.  Rather, they are given a special prompt that is designed to encourage them to specialize in a particular sub-task.  For example, an expert agent that is asked to focus on experiments and evaluation is given a prompt that encourages it to think about the kinds of experiments that it would expect to see in order to support a particular claim, and then compare those hypothesized experiments to the real experiments in the paper.  We found in preliminary testing that giving such instructions to the leader agent tends to work poorly and often ignores details of the instructions, as though the model is overloaded by the number of instructions it is trying to follow, while refactoring the subtask to the separate expert model produces a much higher-quality result.

All agents are given some information about the agent group; they are told how many agents are in the group and the IDs of the agents (while the IDs are not directly used in the communication protocol, they are useful for internal chain-of-thought; for example, the leader might note that it needs to follow up with a particular agent).  In our setting there is always exactly one leader agent, $N$ worker agents for a paper with $N$ chunks, and zero or more expert agents.

\subsection{Communication}

The leader agent is able to send messages to other agents by outputting a special string ("\texttt{SEND MESSAGE:}") followed by the message content.  The message is then broadcast to all other agents in the group.  When an agent receives a message, it is appended to the history as a "system" message with the header "Message from <agent id>:" preceding the message.  The LLM is then run to generate a response to the received message, and this response is always treated as a reply to the leader agent.  Replies from all agents are added to the message history of the leader agent before generating the next output from the leader.

When the leader agent generates an output that does not send a message (and thus does not seek any additional information), the task is complete and we prompt the agent to return the final answer.

\subsubsection*{Error correction}
\label{sec:reviewgen_error_correction}

We attempt to correct a few common errors that occur in practice when agents try to use the communication protocol.  In some cases, agents get stuck in a loop, often when the task is complete.  For example, the leader agent might send a message saying "Thank you all for your feedback and cooperation.", the workers respond with "You're welcome, Agent 0.", the leader says "Thank you all for your responses.", and this loop of unending pleasantries continues.  Such loops typically devolve into exactly the same messages being sent repeatedly, so we check if a message is ever duplicated and if so, we interject with a user message indicating that the message has been duplicated and that it should not be sent again.

We also observe from preliminary experiments that the leader agent does not always remember to follow the protocol for sending a message and simply writes the message body without the necessary header, especially as the conversation grows longer.  This is mitigated by including a short reminder every time the leader agent receives messages, reminding it that it must use the appropriate protocol if it wants to respond.

Finally in some cases the leader agent explicitly addresses a message to one agent (e.g., the expert), but that agent does not recognize the message as being addressed to them.  To identify such cases and speed up inference, we add a prompt instruction with a specific string an agent should output if they wish to not respond to a message.  We detect the presence of any agent ID in a sent message, and if the agent in question outputs the no-response string, we inject a follow-up message reminding them that their name is in the message and encouraging them to respond.

\subsection{Context management}

Our experiments use the 8k-token version of GPT-4.  Versions that can consume more tokens have been developed, but they were not available to us as of this work, and some studies have suggested that even models that can ostensibly consume a large number of tokens may not be able to attend to all of them effectively in practice \citep{qin_nlp_2023,liu_lost_2023}.  {\margfam} can scale beyond the token limit of the base LLM by distributing the input across worker agents, and does not require any individual agent to process a large number of tokens.  However, with a large number of agents or many turns of inter-agent discussion (both of which increase with paper length), the combined tokens of messages sent between agents could cause the message history to eventually exceed the input token limit for especially long papers.

To mitigate the impact of long discussions between agents, we prune old messages from the history on each round of communication.  The pruning strategy is different depending on the agent type.

History length is most limited for the worker agents, which each have a paper chunk occupying most of their token limit, so the histories for workers were trimmed to the initial prompts plus the three most recent messages.  For the leader agent, we observe that (1) a long history is sometimes necessary for in-depth discussions, (2) the majority of tokens in the history arise from all the messages it receives from (potentially many) other agents, and (3) as the leader relays information between other agents, it generally summarizes any important information from messages it receives.  We therefore prune the past messages received from other agents, but keep the full history of outgoing messages.  Finally, for expert agents we never observed issues with the token limit, so no pruning was applied.

\subsection{Review generation}

To tune prompts for review generation, we performed several hundred rounds of manual iteration on a small set of papers from {\dataset} \citep{darcy_aries_2023}.  As the review generation task is somewhat subjective and there are a large number of potential shortcomings with different levels of severity, it is not always straightforward to determine whether a model has made a clear error or if it simply has a difference of opinion with respect to what the most important comments are.  We found it helpful to manually alter some of the papers to create severe and obvious errors that we could expect the model to identify; for example, removing an entire section or adding an unfounded claim (e.g., \reviewtext{the proposed method achieves artificial general intelligence}).  Surprisingly, these "obvious" errors were often not trivial for the system to recognize, making the altered papers useful for finding and mitigating blind-spots.  The final prompts are shown in \autoref{appendix:reviewgen_prompts}, and an outline of our system structure is described in the following paragraphs and shown in \autoref{fig:marg_s_overview}.

We use three independent multi-agent groups to generate different kinds of review comments.  The task prompt given to the leader agent is different for each comment type, and each group has one expert. The comment types are based loosely on points in the ICLR reviewer guidelines.\footnote{\url{https://iclr.cc/Conferences/2023/ReviewerGuide}}  In particular, it asks \reviewtext{\textcolor{black}{[...]} is the submission clear, technically correct, experimentally rigorous, reproducible, does it present novel findings (e.g. theoretically, algorithmically, etc.)?}  We group and slightly reframe these points to arrive at the following comment types:

\subsubsection*{\textbf{Experiments and evaluation:}} The leader is instructed to focus on verifying that the experiments and theoretical proofs are correct and adequately support the paper's claims.  The expert in this group is told to "design high-quality experiments" given the main claims made in the paper, inspired in part by the fact that making predictions is an effective active reading strategy to improve comprehension in humans \citep{fielding_how_1990,duke_effective_2009}.  In preliminary experiments without the expert, the model could identify some bad experiments and give generic comments, but struggled to realize when an experiment was missing.  Explicitly designing experiments provides a baseline with which to compare the experiments in the paper, allowing the model to recognize missing or incomplete experiments.

\subsubsection*{\textbf{Clarity and reproducibility:}} The leader is instructed to focus on ensuring that the paper clearly explains key concepts and proposed methods, and that it provides all necessary details to implement any proposed methods and reproduce experiments.  The expert in this case is instructed to be "highly curious" and to ask questions of the leader agent in order to learn more about the paper.  This process aids in identifying any questions that {\em can't} be answered based on the paper, which become comments.

\subsubsection*{\textbf{Novelty and impact:}} The leader is instructed to focus on the novelty and impact of the paper.  However, we note that for our study the task of accurately retrieving related work is out of scope, so this comment type is limited to identifying errors in the paper's own explanations.  Specifically, the model is instructed to verify that the paper clearly states and justifies its motivations, goals, and key findings, and that it thoroughly discusses how it fits into the existing literature.  The expert in this case is instructed to be skeptical of the paper and ask questions to determine if it actually makes a significant contribution to its field.

\subsection{Refinement}

After generating a review, we find that it is very helpful to include a "refinement" stage, in which the model is given the review comments and asked to improve (or remove) them.  Various errors can arise during the initial comment generation, and we observe that models tend to be poor at self-reflection and correction during that stage.  Including refinement as a separate stage can resolve many of the errors introduced during the initial generation.

To refine comments, we initialize a new multi-agent group with no expert agent.  For each comment, we provide the comment to the leader agent with a prompt instructing it to ensure that the comment is clear, that it is specific, and that it is valid (i.e., does not suggest something that is already done in the paper).  The model outputs a list; usually this list contains one element (the newly-refined comment), but may contain more (if the original comment mixed two different suggestions) or be null (if the comment was invalid).  The comments are processed independently (i.e., by separate multi-agent groups).

\section{Baseline methods}

In this section, we will describe the baseline methods that we compare against our multi-agent approach.  We consider three baselines: a simple baseline that treats chunks independently and uses a one-line prompt, a baseline that treats paper chunks independently but uses a more sophisticated prompt, and a recently proposed method for generating peer-review feedback \citep{liang_can_2023}.  Prompts for these methods can be found in \autoref{appendix:reviewgen_prompts}.

\subsection{Single-Agent Review Generation with Basic prompt (SARG-B)}

This baseline is designed to emulate a simple approach that a ChatGPT user might use to get feedback on their paper if they did not want to do any prompt tuning.  We use a single agent to generate all comments for the paper.  The paper is split into the same chunks as for the multi-agent baseline, but the chunks are processed independently using a very simple prompt:

\begin{displayquote}
    \begin{singlespace}
        \begin{lstlisting}[basicstyle=\ttfamily\tiny,breaklines=true,breakindent=0pt]
Write feedback comments in the style of a scientific paper review for the following portion of a scientific paper.  You can skip minor grammar comments.
        \end{lstlisting}
    \end{singlespace}
\end{displayquote}

After applying the model to each chunk, the resulting comment lists are combined by a similarly simple prompt:

\begin{displayquote}
    \vspace{-0.2\baselineskip}
    \begin{singlespace}
        \begin{lstlisting}[basicstyle=\ttfamily\tiny,breaklines=true,breakindent=0pt]
Here are some lists of review comments that were made about different portions of the paper: <comment lists>
Merge these lists into a final list of review comments.  Any comments that are duplicates (saying essentially the same thing as other comments) should be merged or deleted.
        \end{lstlisting}
    \end{singlespace}
    \vspace{-0.7\baselineskip}
\end{displayquote}

\subsection{Single-Agent Review Generation with Tuned Prompt (SARG-TP)}

This baseline is designed to emulate a more sophisticated approach that a ChatGPT user might use to get feedback on their paper if they were willing to do some prompt tuning.  We use a single agent to generate all comments for the paper, but we use a more sophisticated prompt (\autoref{appendix:reviewgen_prompts_sarg_tp}) that is designed to encourage the model to generate more specific and actionable comments.  As with the other simple baseline, we generate comments independently for each paper chunk and then merge the resulting lists with GPT.

Similarly to our multi-agent method, we include a refinement step in this baseline.  For each paper chunk, we give the model the chunk and the final list of comments, and ask it to output a new, refined list of comments.  This provides an opportunity to remove incorrect comments that arise from the independent processing of each chunk.  For example, if one chunk contains the introduction but not the experiments, the model might initially write a comment that claims the experiments are missing, but in the refinement stage will be able to prune it when it sees the chunk that does contain experiments.

\subsection{Multi-Agent Review Generation with Tuned Prompt ({\margfam}-TP)}

This baseline is designed to provide a direct comparison with the prompt-tuned single-agent baseline and explore the benefits of multi-agent modeling.  Whereas our full {\margmain} approach leverages several advantages of multi-agent that would be difficult to directly compare in a single-agent setting (e.g., the use of expert agents), this multi-agent baseline uses a prompt designed to be as similar as possible to the prompt-tuned single-agent baseline.  Of course, we still must include some instructions that explain the communication protocol and instruct the agents to work together, but the task prompt includes all the same language as in the single-agent setting.  Similarly, we use a refinement prompt that is as similar as possible to the single-agent setting, although the refinement stage still differs in that we do not manually apply it on each chunk (as this would defeat the point of using multiple agents).

\subsection{Multi-Agent Review Generation with Specialized Agents ({\margmain})}

Our full {\margmain} approach is described in \autoref{sec:reviewgen_multi_agent}, and uses three independent multi-agent groups to generate different kinds of review comments.  {\margmain} outputs the concatenation of the three mini-reviews generated by those groups. In addition to the full approach, we evaluate each of the three mini-reviews separately.  We refer to these as {\margmain} (experiments), {\margmain} (clarity), and {\margmain} (impact).  In addition, we include a "no refinement" baseline that skips the refinement stage.

\subsection{\citet{liang_can_2023} baseline (LiZCa)}

We also compare against a recently proposed method for generating peer-review feedback \citep{liang_can_2023}, which we refer to as "LiZCa" (from the names of the lead authors of that paper; the method was not given a name in that work).  Unlike our methods, this method simply truncates the paper rather than applying to multiple chunks.  In addition, it includes the captions of figures and tables in the input.

The prompt used in \citet{liang_can_2023} instructs the model to generate an "outline" style review, and includes non-actionable positive comments.  Fortunately, when comparing their method's comments with real reviews, they developed a prompt to extract and merge the parts of an outline that focus on "criticisms" and to ignore minor grammar comments.  This roughly matches the type of comments we target, so we use that prompt to produce the final list of comments that we use in this baseline.

We note that \citet{liang_can_2023} used a different PDF parsing library (pikepdf) than ours (Grobid), but for consistency with our other baselines we run it with Grobid.

\section{Automated evaluation}

To automatically evaluate the quality of generated reviews, we measure their overlap with real reviews from papers in the {\dataset} corpus \citep{darcy_aries_2023}.  That is, we attempt to match the generated comments to comments extracted from real (human-written) reviews.  Because {\dataset} only has comment annotations for a small set of reviews, we use GPT\footnote{We use \texttt{gpt-4-0613}, which has an 8192-token capacity; larger models have been developed but were not available to us while conducting this work.} to extract comments from all reviews for a subset of 30 papers and treat this as our test set.  To match our intended type of feedback, GPT is instructed to focus only on actionable feedback comments and to ignore minor comments on style and grammar.

We note that this form of evaluation is imperfect in that real reviewers do not always identify every reasonable critique of a paper, and in some cases they may make critiques that are unreasonable.  Thus, the generated review could contain good comments that happen to be different from ones the real reviewers made, or it could miss comments that are actually invalid.  Thus, the measured overlap should be treated as a lower bound for the fraction of good-quality comments.  In addition, the nuanced nature of the matching task makes it impossible to fully capture the similarities and differences between real and generated comments using binary alignments, and this could lead to biases.  We nonetheless use automated evaluation as an inexpensive but rough approximation of the relative quality of different methods, and separately conduct a user study in \autoref{sec:reviewgen_user_study} to obtain a more realistic evaluation.

The matching procedure and results are outlined in the following subsections.

\subsection{Measuring overlap}

\label{sec:aligning_comment_lists}

\begin{table*}
    \centering
    \begin{tabular}{p{0.4\textwidth}|p{0.58\textwidth}}
        \toprule
        Real-reviewer comment & Generated comment  \\
        \midrule
        \textbf{The experimental methodology used in the paper is not well detailed}, making it difficult to reproduce the reported results. & More details about the experiments conducted would be beneficial. This should \textbf{include information about the datasets used, the training process, and the evaluation process}. To ensure the reproducibility of the results, \textbf{consider providing the code used to implement the model, the specific parameters used, and any other necessary information}. This will allow other researchers to replicate your work and further validate your findings.\newline \textbf{[high relatedness, more specific]} \\
        \midrule
        The paper does not include enough baselines for Fair Federated Learning to compare against. Even if some methods do not satisfy privacy considerations, they should still be included for the reader to understand how the proposed method compares against such methods, especially given that the results are not promising. \textbf{Some baselines to consider include Cui et al or Tran et al.} & The authors should \textbf{consider including a comparison of their proposed method with existing methods} in the experimental results section. This would help to highlight the advantages and improvements of their proposed method.\newline \textbf{[high relatedness, less specific]} \\
        \midrule
        \textbf{The datasets used in the study are not representative} due to their simplicity and experimental nature. & The evaluation of the proposed method may not be comprehensive enough. The authors could \textbf{include more datasets} in their evaluation to demonstrate the robustness of their method. The paper could benefit from a more detailed discussion on the limitations of the proposed method.\newline \textbf{[medium relatedness, more specific]} \\
        \bottomrule
    \end{tabular}
    \vspace{\baselineskip}
    
    \caption{Aligned pairs of comments with corresponding relatedness and relative specificity scores from the alignment model; the bold is added to emphasize key differences.  Notice that in the third row with "medium" relatedness, the reviewer comment is suggesting that the datasets need to be more representative (but a larger number of datasets is not necessarily needed) whereas the generated comment only asks for more datasets (not identifying the issue with the current datasets).  In the two "high" relatedness cases, one comment fully subsumes the other (high relatedness) but includes much more specific details and rationales (less/more relative specificity).}
    \label{tab:reviewgen_alignment_example}
\end{table*}

Given a set of generated review comments $C_{gen}$ and the set of ground truth real-reviewer comments $C_{real}$ for a paper, we automatically align individual comments between the reviews that have the same meaning.  That is, we ultimately obtain a binary label for every comment pair $(C^i_{gen}, C^j_{real})$ indicating whether the two comments are making the same request.  To do this, we begin with a "many-many" matching stage that efficiently compares the full set of comments in both reviews and identifies possibly-matching pairs, followed by a more accurate (but more expensive) pairwise stage that examines the candidate pairs to produce a final list.

In the many-many matching stage, we feed all comments from both reviews into GPT-4 and prompt it to output a list of all matching comments.  As GPT has somewhat inconsistent performance, we do five such passes, randomly permuting both the order of comments within each review and the order in which reviews are presented.  The final output of this stage is the list of comment pairs that were produced by at least two of the five runs---a ratio we heuristically found to work well in preliminary experiments.

In the pairwise stage, we give one comment pair at a time to GPT and prompt it to produce two scores: one of four levels of relatedness ("none", "weak", "medium", or "high"), and a "relative specificity" ("less", "same", "more") indicating how specific the generated comment is relative to the real review comment.  To be considered a match, a comment pair must have "medium" or "high" relatedness, and the generated comment must have "same" or "more" specificity compared to the human comment.  An example of an aligned pair of comments can be found in \autoref{tab:reviewgen_alignment_example}.

The final output is a list of alignment edges between the lists of generated and real-reviewer comments.  We note that this may result in a many-many mapping; one generated comment might match multiple reviewer comments, and one reviewer comment might match multiple generated comments.  This can happen when there are similar comments within one list or if, for example, a reviewer makes a broad suggestion like "Evaluate on more datasets" and the generated review contains several comments, each with a different specific dataset recommendation.

\subsection{Metrics}
\newcommand{\leftintersect}{\overset{\leftarrow}{\cap}}
\newcommand{\rightintersect}{\overset{\rightarrow}{\cap}}

Using the alignments between $C_{gen}$ and $C_{real}$, we evaluate several metrics, described below.  However, we note that the many-many nature of the mapping between the comments indicates that these are not proper sets, and traditional set-based metrics such as the union and intersection are not well-defined.  For our purposes, we define directional intersection operators $\leftintersect$ and $\rightintersect$ representing the set of aligned elements in the left or right operand, respectively.  For example, $C_{gen} \leftintersect C_{real}$ is the set of elements of $C_{gen}$ that align to any element in $C_{real}$.

\begin{itemize}
    \item \textbf{Recall:} $\frac{|C_{gen} \rightintersect C_{real}|}{|C_{real}|}$, the fraction of real-reviewer comments that are aligned to any generated comment.
    \item \textbf{Precision:} $\frac{|C_{gen} \leftintersect C_{real}|}{|C_{gen}|}$, the fraction of generated comments that are aligned to any real-reviewer comment.
    \item \textbf{(Pseudo-)Jaccard:} The Jaccard index is a commonly-used measure of set overlap.  Let $intersection = \frac{|C_{gen} \leftintersect C_{real}| + |C_{gen} \rightintersect C_{real}|}{2}$; then the Jaccard index is $\frac{intersection}{|C_{gen}| + |C_{real}| - intersection}$.   
\end{itemize}

To compute these metrics over a set of papers, we macro-average on the level of reviews.  That is, given a set of papers in our test set, we generate a review for each, measure the aforementioned metrics between each generated review and each corresponding real review, and then average all of the results to obtain a single value for each metric.

\subsection{Results}

\begin{table*}[t]
    \centering
    \begin{tabular}{l|cccc}
        \toprule
        Method                         & Recall          & Precision       & Jaccard         & \# comments \\
        \midrule
        SARG-B                         &   7.43          &     1.40        &   1.25          &       19.7  \\
        SARG-TP                        &  10.62          &     4.61        &   3.46          &       11.6  \\
        {\margfam}-TP                  &   8.49          &     5.34        &   3.52          &        8.5  \\
        LiZCa                          &   9.67          &  \textbf{9.96}  &  \textbf{5.58}  &        4.0  \\
        \midrule
        {\margmain}                    &  \textbf{15.84} &     4.41        &   3.53          &       19.8  \\
        \hspace{1em} no refinement     &  11.92          &     3.32        &   2.70          &       18.3  \\
        \hspace{1em} experiments-only  &   4.36          &     4.83        &   2.23          &        4.1  \\
        \hspace{1em} clarity-only      &   3.25          &     2.65        &   1.46          &        6.9  \\
        \hspace{1em} impact-only       &   8.88          &     4.75        &   3.32          &        8.8  \\
        \midrule
        Human                          &   9.42          &    12.00        &   5.45          &        4.7  \\
        \bottomrule
    \end{tabular}
    \vspace{\baselineskip}
    \caption{Automated evaluation results with recall, precision, and Jaccard values, in addition to the average number of comments generated by each method.  The proposed {\margmain} method outperforms all baselines in terms of recall, but generates more comments than other baselines and thus has lower precision and Jaccard scores.}
    \label{tab:reviewgen_automated_results}
\end{table*}

\begin{table*}
    \centering
    \begin{tabular}{p{0.12\textwidth}|p{0.85\textwidth}}
        \toprule
        Method & Example comment \\
        \midrule
        SARG-B & The paper could benefit from a more detailed discussion of the results, including the implications of the findings and how they contribute to the existing body of knowledge. \\
        \midrule
        LiZCa & The experimental evaluation could be more comprehensive. The authors should consider including more diverse tasks and environments in their experiments to demonstrate the robustness of their method. The paper could benefit from a more detailed analysis of the experimental results, including a discussion on why the proposed method outperforms the baselines. \\
        
        \midrule
        {\margmain} (experiments) & The authors have compared their method with several baselines, including DeepMDP, HiP-BMDP-nobisim, Distral, PCGrad, GradNorm, and PEARL. However, it would be beneficial to include comparisons with other state-of-the-art methods in multi-task and Meta-RL setups to further validate the effectiveness of the proposed method. This would help ensure that the results are not specific to the current set of comparisons and can generalize across different settings. Additionally, providing a detailed discussion on why the proposed method outperforms each baseline could offer more insights into the strengths and weaknesses of the proposed method. \\
        \midrule
        {\margmain} (clarity) & The paper mentions an encoder that maps observations from state space to a learned, latent representation, but it does not provide specific details about the type of encoder used or the process of how it learns the latent representation. These details are crucial for understanding how the model works and how it achieves its performance. Therefore, I recommend that the authors include this information in the paper. \\
        \midrule
        {\margmain} (impact) & The authors provide a theoretical proof for the 'Transfer bound' formula, which is a significant contribution. This formula is crucial for measuring the transferability of a policy learned on one task to another, taking into account the error from the learned representation. However, to ensure its robustness and applicability in real-world scenarios, it would be beneficial if the authors could empirically test this formula in reinforcement learning environments. For instance, the agent could be trained on one game and then tested on a different game with similar mechanics. This would provide empirical evidence supporting the theoretical proof and demonstrate the practical utility of the formula. \\
        \bottomrule
    \end{tabular}
    \vspace{\baselineskip}
        \caption{Example comments generated by each method (SARG-TP and {\margfam}-TP omitted for brevity) for the same paper.  Qualitatively, we find that {\margmain} writes relatively long and specific comments, whereas other methods tend to write shorter and more generic comments.}
    \label{tab:reviewgen_example_comments}
\end{table*}

We include a selection of example generated comments in \autoref{tab:reviewgen_example_comments}.  Results of the automated evaluation are shown in \autoref{tab:reviewgen_automated_results}.  We additionally include a human-review baseline, which is the average of the metrics computed between each real review and each other real review for the same paper (i.e., $\frac{1}{n}\sum^n_{i=1}\text{metric}(human_i, \{human_k | k \ne i\})$.  Note that while this is theoretically unbiased for recall, it may result in lower precision and Jaccard scores for human reviewers.

We find that our proposed {\margmain} method outperforms all baselines in terms of recall, but generates more comments than other baselines and thus has lower precision and Jaccard scores.  With that said, we believe that recall is the most important metric in this evaluation.  While higher precision and Jaccard should be preferred at similar levels of recall, it is relatively easy for a human to recognize and ignore bad comments; thus, it is more important for the system to maximize the number of good comments than to minimize the number of bad ones.

The simple baseline (SARG-B) performs poorly on all metrics; despite being tied with {\margmain} for the highest number of generated comments, it has the lowest recall of all methods.  This is not unexpected, but highlights the importance of careful prompting with GPT-4.

Interestingly, we find that between SARG-TP and {\margfam}-TP (which use essentially the same task prompt), SARG-TP generates more comments and has better recall.  This suggests that simply applying a multi-agent approach does not always result in a performance improvement; instead, the use of multiple agents enables the design of richer internal problem-solving structures via expert agents.  Indeed, we see that the specialized {\margmain} (impact) is able to approximately match the performance of {\margfam}-TP despite focusing on only one type of comment.

We notice that the human baseline actually has a lower recall than some of the LLM baselines, although it has the highest precision.  This is consistent with the results of \citet{liang_can_2023}, which found that Human-Human agreement was slightly lower than LiZCa-Human agreement.\footnote{Note that while the relative differences are similar to those reported in \citet{liang_can_2023}, our absolute recall scores are lower.  We conjecture that this is primarily due to differences in the alignment step; in particular, the pairwise filtering makes our approach more conservative.}  Humans generate fewer comments than other approaches, which offers a partial explanation for the low recall, but it is nonetheless interesting to observe that human reviewers can have very different perspectives of the same work, and highlights the challenge of the review generation task (and the potential weaknesses of alignment-based evaluation).

\subsubsection*{{\margmain} ablations:}

    Among the sub-reviewers of {\margmain}, the impact-focused model tends to produce the best results.  The experiment-focused model does well considering the small number of comments it produces, but as it produces half as many comments as the impact model it also has half the recall.  Finally, the clarity-focused model struggles compared to the other two.  The poor performance of the clarity model may be due in part to the subjective nature of clarity judgements and the fact that language models do not necessarily perceive text in the same way that humans do (e.g., humans prefer that terms be defined before they are used, but a model that consumes a full document at once might not see a problem if terms are defined later).  In addition, we note that due to the fact that the input does not capture visual information such as figures, tables, and the arrangement of symbols in equations, there are many resulting clarity issues that are not present in the full paper, and getting the model to identify the "real" issues from among the large number of parsing- and input-format-related issues is challenging.

We observe that without the refinement stage, {\margmain}'s performance is reduced on all metrics, but it still obtains reasonable results; recall remains the second-highest of all methods.  Interestingly, the number of generated comments is slightly lower than with the refinement stage, indicating that the refinement stage splits one comment into multiple comments more often than it prunes comments.

\subsubsection*{Effect of the matching thresholds}

\begin{figure}[ht]
    \centering
    \begin{minipage}{0.45\textwidth}
        \centering
        \includegraphics[width=\textwidth]{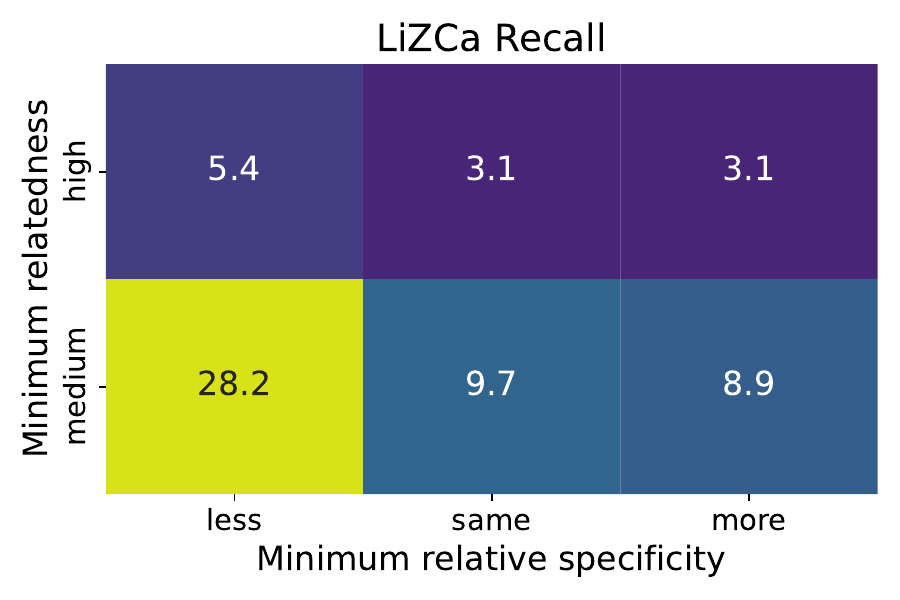}
    \end{minipage}\hfill
    \begin{minipage}{0.45\textwidth}
        \centering
        \includegraphics[width=\textwidth]{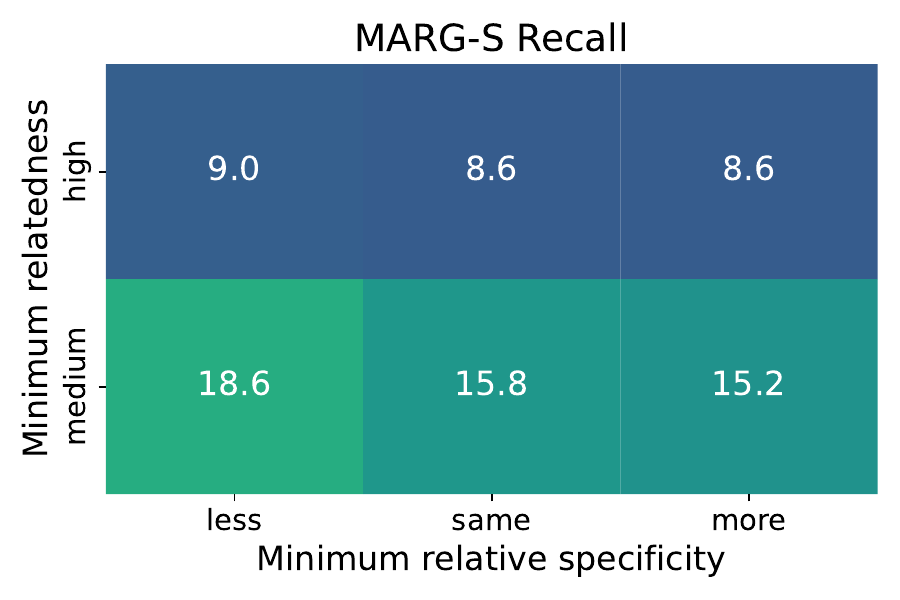}
    \end{minipage}
    \caption{Recall of {\margmain} and LiZCa for different alignment cutoff levels of relatedness and relative specificity.  The ("medium", "same") cell corresponds to our default setting.  LiZCa obtains very high recall in the most lenient setting, but rapidly drops for stricter settings that prevent vague comments from being counted as matches.  {\margmain} obtains relatively consistent results for all levels of specificity (as most of its comments are considered "more" specific) but still experiences a decline when requiring highly-related matches.}
    \label{fig:reviewgen_matching_thresholds}
\end{figure}

We qualitatively observe that several of the methods we evaluate produce many generic or vague comments.  Many of these are not counted towards recall due to our constraint that a generated comment must be equally or more specific compared to the real comment it matches with.  In addition, some aligned pairs of comments are questionable, especially for pairs that have only "medium" relatedness; for example, a comment asking for a "more thorough comparison" to baselines is considered a match for one that indicates that the proposed method underperforms the baselines in some cases.

To evaluate the impact of threshold choices, we select our method and the LiZCa baseline and evaluate all combinations of thresholds for relative specificity and "medium" or "high" relatedness.  The results are shown in \autoref{fig:reviewgen_matching_thresholds}.

The difference between thresholds is striking in the figure.  LiZCa generates a large number of comments that broadly match to a real comment, but are much more vague (and thus less useful) and often do not have exactly the same meaning.  When either the relatedness or the specificity thresholds are increased even by one step, the recall drops sharply.  In contrast, {\margmain} loses relatively little recall at higher specificity thresholds, as it almost always outputs specific comments.  However, {\margmain} still has a drop when requiring "high" relatedness, indicating that it has many matches in which the generated comment does not express exactly the same sentiment as the real one.

\subsubsection*{Cost}

\begin{table}
    \centering
    \begin{tabular}{l|rr}
        \toprule
        Method                          & Input tokens & Generated tokens \\
        \midrule                        
        SARG-B                          &     15,532 &   3,249 \\
        SARG-TP                         &     54,914 &   6,853 \\
        {\margfam}-TP                   &    134,860 &   5,264 \\
        LiZCa                           &      7,398 &     869 \\
        {\margmain}                     &  1,236,344 &  51,255 \\
        \bottomrule
    \end{tabular}
    \vspace{\baselineskip}
    \caption{Average number of input and generated tokens per paper for each method.  This includes tokens used for internal discussion in multi-agent methods, but not tokens used outside of the method (e.g., for measuring the alignment metric).  {\margmain} generates substantially more tokens than other methods, and thus is more expensive to run.}
    \label{tab:reviewgen_method_cost}
\end{table}

In \autoref{tab:reviewgen_method_cost} we report the average number of tokens generated by each method during the alignment-based evaluation.  LiZCa generates the fewest tokens and has the best cost to recall ratio overall, making it an attractive choice in budget-constrained settings.  While {\margmain} has the best recall, it also generates roughly an order of magnitude more tokens than other methods, suggesting that it takes on diminishing returns in efficiency to obtain the recall improvement.

The extra tokens used by {\margmain} result in it taking roughly an hour longer than other methods to generate reviews.  This may serve as an inconvenience in practice, and it would be beneficial to explore ways to reduce it.  For example, it may be possible to dynamically switch to cheaper LLMs to handle simpler messages or develop methods to route communications more effectively (reducing the number of redundant messages).  We also note that our implementation performs only one inference at a time for simplicity, but in theory, it is highly parallelizable (due to having three separate groups for different comment types, separate groups for the refinement stage, and several agents communicating at once in each group), and the time needed to generate a review could likely be reduced by 2-10x depending on the document size.

\section{User study}
\label{sec:reviewgen_user_study}

We conduct a user study to obtain a more reliable (but more expensive) evaluation compared to the automated metrics.  To reduce burden on participants, we only evaluate a subset of methods in the user study: {\margmain} (our best method on the automated metrics), LiZCa (baseline from prior work), and SARG-B (the simplest baseline).

\subsection{Study design}

\subsubsection*{Participants}

We recruit 9 volunteers\footnote{While this is a small number of participants, we note that the number of rated comments is much higher (each participant rates many comments per method), and we obtain statistically significant conclusions from mixed-effect analyses in which we control for participant bias as a random effect.} from a large research organization to participate in the study.  All participants are researchers in the fields of natural language processing and human-computer interaction.

\subsubsection*{Survey}

\begin{figure*}[ht]
    \centering
    \includegraphics[width=0.6\textwidth]{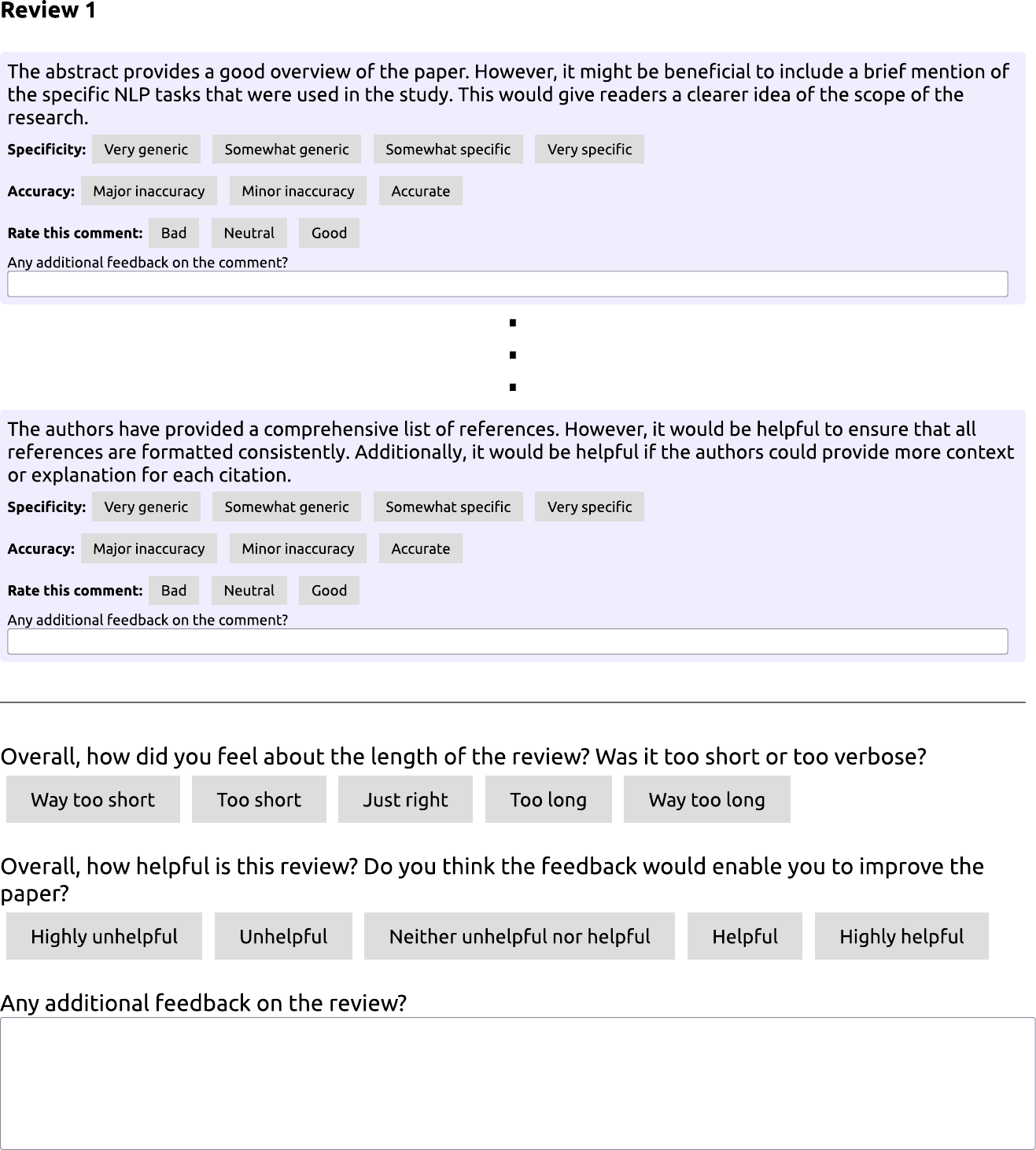}
    \caption{The survey interface.  Participants were asked to rate the specificity, accuracy, and overall helpfulness of each comment, and to rate the overall review.}
    \label{fig:reviewgen_survey_interface}
\end{figure*}

The study was conducted using a web interface in which participants could upload a paper PDF.  We then ran each review generation method to produce a set of reviews, where each review was a list of comments.  When all reviews were generated, participants would receive an email notification with a link to page with reviews and a set of survey questions, depicted in \autoref{fig:reviewgen_survey_interface}.  The survey page did not describe the review generation methods or give any indication of which method generated a given review, and the generated reviews were displayed in a random order to reduce bias (the order of comments within reviews was not randomized, however).

For each comment, participants were asked to rate its specificity, accuracy, and to provide an overall rating.  The following guidelines for these ratings were provided at the start of the survey:
\begin{itemize}
    \item \textbf{Specificity:} Does the comment make a suggestion specific to the paper, or is it generic (could apply to many papers)?  Please note that a comment may be verbose without being specific, or vice versa.
    \item \textbf{Accuracy:} Does the comment display an accurate understanding of the paper and make a valid critique?  For example, suppose a comment says the paper is missing statistical significance tests and should include them.  If the paper doesn't have significance tests and could potentially benefit from including them, please rate the comment as "accurate" (even if the importance of those tests is questionable).  If the paper has tests on one or two results but not all, and the comment doesn't mention this, the comment would have a "minor" inaccuracy.  If the paper already has extensive significance tests or provides substantial justification for not including them, the comment would have a "major" inaccuracy.
    \item \textbf{Overall rating:} How helpful is the comment overall?  Is the comment one that you would want to see in a review (Good), one that you might not mind seeing but don't care much about (Neutral), or one that is useless or invalid (Bad)?
\end{itemize}

In addition, participants were asked questions at the end of each review.  Specifically, they were asked to rate whether the review was too long or too short on a 5-point scale and to provide an overall rating for the review on a 5-point scale.

Finally, we asked participants about their research and reviewing experience, and about their authorship of the submitted paper.

\subsection{Total good comments}

\begin{table}
    \centering
    \begin{tabular}{l|cccc}
        \toprule
        Method      & Bad & Neutral & Good & Total \\
        \midrule
                   &   Bad & Neutral &  Good          & Total   \\
        SARG-B     &  10.2 &     4.2 &   1.7          &  16.1   \\
        LiZCa      &   2.7 &     0.9 &   0.3          &   3.9   \\
        {\margmain}&   8.1 &     5.3 &   \textbf{3.7} &  17.1   \\
        \bottomrule
    \end{tabular}
    \vspace{\baselineskip}
    \caption{Average number of each comment rating per review for each method.  {\margmain} generates the most good comments.  LiZCa generates substantially fewer comments than the other methods, and therefore has the fewest bad comments per review but also the fewest good comments.}
    \label{table:reviewgen_user_study_rating_totals}
\end{table}

As in the alignment-based evaluation, we argue that bad comments have relatively small cost compared to the value of good comments.  As there is no straightforward way to adjust the total number of generated comments (unlike in a classification task, where the decision threshold could be adjusted continuously), the total number of good comments is the most appropriate metric with which to compare methods.

\autoref{table:reviewgen_user_study_rating_totals} shows the average number of each comment rating per review for each method.  We find that {\margmain} generates more good comments than SARG-B (p=0.09, related-sample t-test) and LiZCa (p=0.003).  LiZCa generates substantially fewer comments than the other methods, and therefore has the fewest bad comments per review but also the fewest good comments.

\subsubsection*{Review length}

{\margmain} generates the most good comments, but does this come at the cost of generating overly-long reviews?  It seems that in general, {\margmain} reviews do tend to be longer than authors would like, while LiZCa reviews are too short.  Specifically, {\margmain} was rated as "way too long" by 6 of the 9 participants (and "just right" by the other three), while LiZCa was rated as "too short" by 4, "way too short" by 3, and "just right" by 2 of the participants.  SARG-B occupied a middle ground, rated as "too short" by 2, "too long" by 3, and "just right" by 4 of the participants.  Although SARG-B generates a similar number of comments as {\margmain}, the comments it generates are much shorter, which is likely why its length is perceived as being more reasonable.

\subsection{Average comment ratings}

\begin{figure*}
    \centering
    \includegraphics[width=0.9\textwidth]{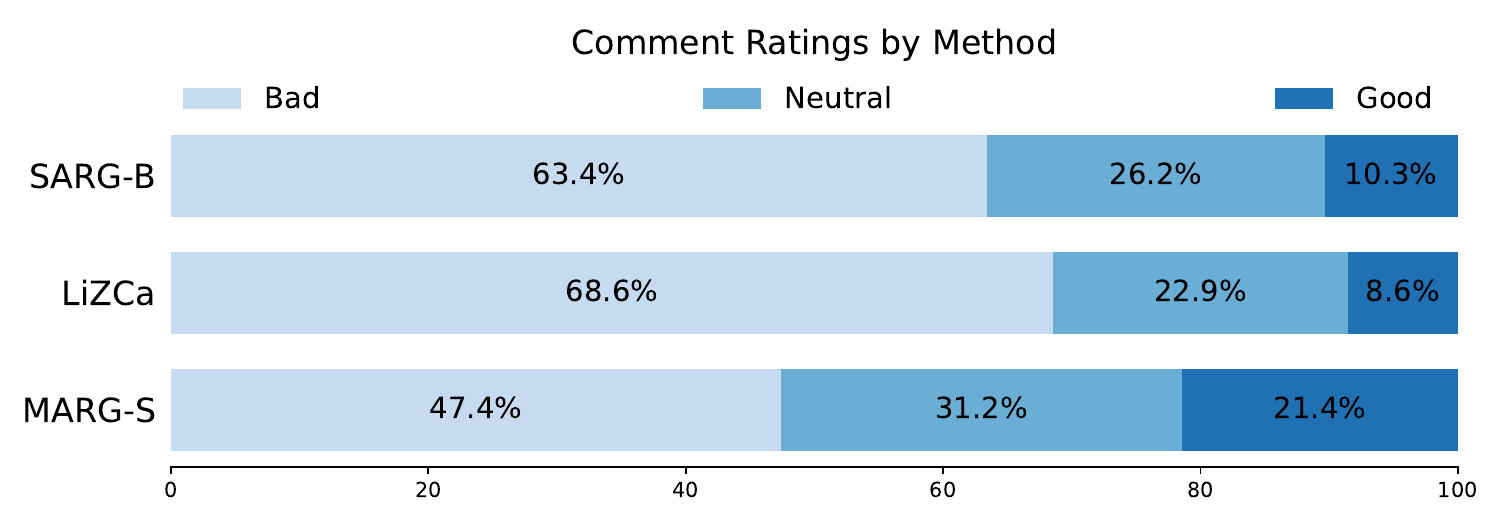}
    \caption{Average quality ratings for each method.  LiZCa and SARG-B are rated similarly, while {\margmain} has over twice the fraction of "good" comments compared to the other two methods.}
    \label{fig:reviewgen_user_study_ratings}
\end{figure*}
\begin{figure*}
    \centering
    \includegraphics[width=0.9\textwidth]{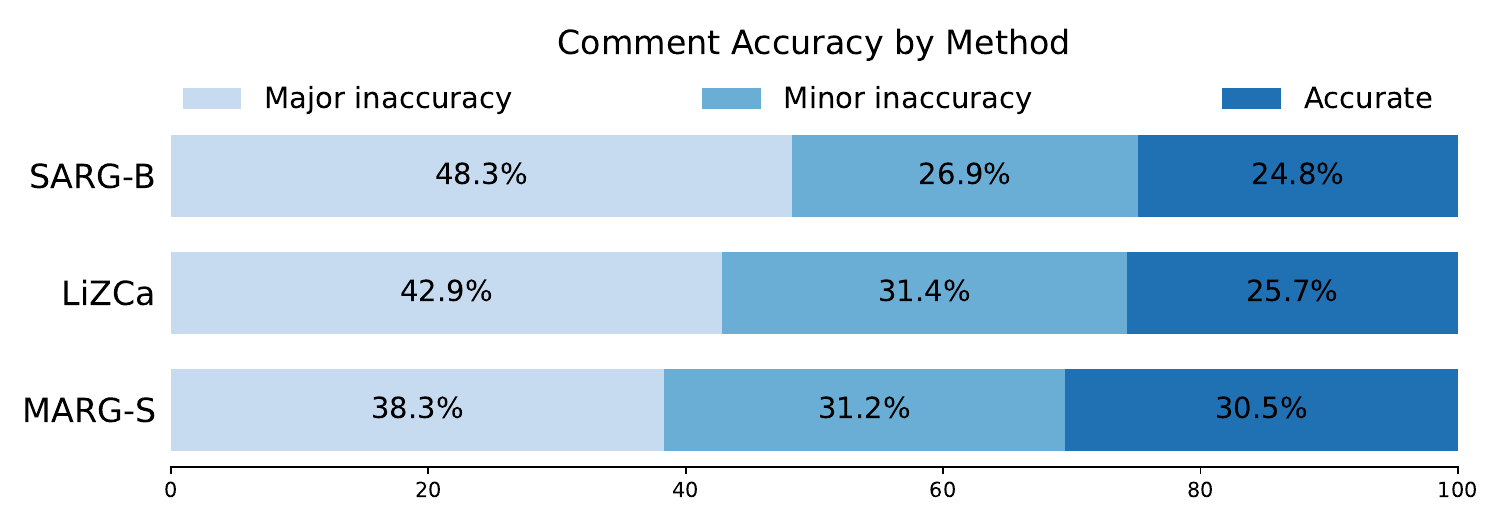}
    \caption{Average accuracy ratings for each method.  {\margmain} has the most fully accurate comments by a small margin, and SARG-B has the most major inaccuracies, but all methods have similar accuracy distributions overall.}
    \label{fig:reviewgen_user_study_accuracy}
\end{figure*}
\begin{figure*}
    \centering
    \includegraphics[width=0.9\textwidth]{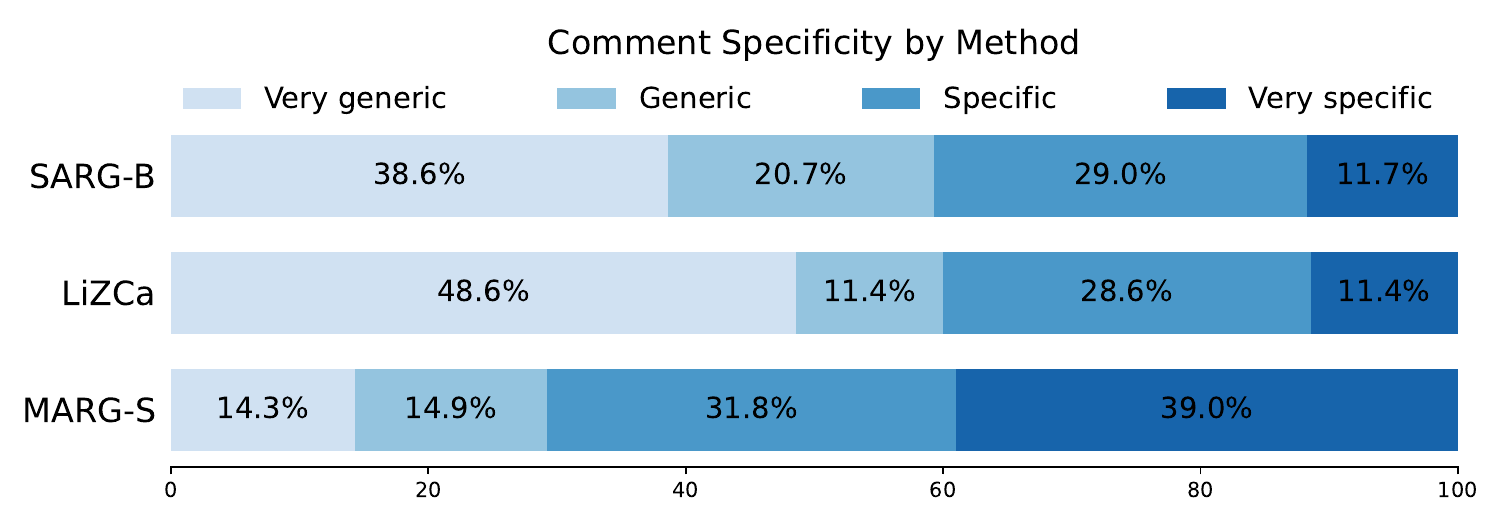}
    \caption{Average specificity ratings for each method.  LiZCa and SARG-B have similar proportions of the specific and very-specific comments, but LiZCa has more very generic comments.  {\margmain} is extremely specific compared to the other two methods; 71\% of {\margmain} comments are rated specific or very specific, compared to only 40\% for LiZCa and SARG-B.}
    \label{fig:reviewgen_user_study_specificity}
\end{figure*}

The distribution of user ratings of comment quality, accuracy, and specificity are shown in \autoref{fig:reviewgen_user_study_ratings}, \autoref{fig:reviewgen_user_study_accuracy}, and \autoref{fig:reviewgen_user_study_specificity}, respectively.

We find that {\margmain} has the highest proportion of "good" comments, and is significantly better than SARG-B (p=0.02 for per-comment Barnard's exact test, p=0.12 for per-user related sample t-test), although the difference between {\margmain} and LiZCa is not significant (p=0.09 per-comment, p=0.16 per-user).  When asked about the overall helpfulness of the reviews, participants rated {\margmain} as an average of 1.0 points higher on the 5-point scale (and 5 of the 9 participants rated {\margmain} as 2-3 points higher).

The accuracy ratings in \autoref{fig:reviewgen_user_study_accuracy} show a similar trend as the comment quality ratings.  {\margmain} has the highest proportion of fully accurate comments, but the differences are not significant.  

The most striking difference between the methods is in specificity.  {\margmain} has "very specific" comments at more than triple the rate of the other two methods, a significant increase (p=0.002, per-user related-sample t-test).  Overall, 71\% of its comments are rated as "specific" or "very specific", compared to only 40\% for LiZCa (p=0.08).

Finally, we observe that {\margmain} has a high proportion of "good" comments rated by users despite having a relatively low precision in the automated evaluation (\autoref{tab:reviewgen_automated_results}).  The difference suggests that it may generate many comments which are helpful but also different than the kinds of suggestions a real reviewer would tend to make.  This could be a promising sign indicating that {\margmain} can serve as a useful source of novel inspiration for authors---even when the paper has already been reviewed by humans---and that it may be a source of inspiration for reviewers as well.\footnote{Of course, it is also possible that the kinds of novel suggestions {\margmain} makes only {\em appear} useful to authors and actually do not improve the paper in ways that reviewers (or readers) care about.  Measuring the extent to which this is the case would require a much more sophisticated study of how these comments affect the long-term impact of papers, and we leave this to future work.}

\subsection{Relationships between factors}

\begin{table*}
    \centering
    \begin{tabular}{lrrrr}
        \toprule
                                       {} &  Coef &  Std. err &  z-value &  Pr(>|z|)       \\
        \midrule

        {\margmain}                    &  0.19 &      0.33 &     0.58 &        0.56 \\
        LiZCa                          & -0.66 &      0.52 &    -1.28 &        0.20 \\
        Inaccuracy (minor)             & -1.33 &      0.35 &    -3.83 &   $10^{-3}$ \\
        Inaccuracy (major)             & -5.54 &      0.58 &    -9.62 &  $10^{-21}$ \\
        Specificity (specific)         &  1.90 &      0.36 &     5.33 &   $10^{-7}$ \\
        \midrule
        Random effect std. dev            & \multicolumn{4}{c}{$\sigma$ = 0.92}            \\
        \bottomrule
    \end{tabular}
    \vspace{\baselineskip}
    \caption{Cumulative link fixed effects for specificity, accuracy, and method on the overall rating of a comment.  Specificity is positively associated ratings, as is accuracy (inaccuracies have a negative effect).  The review generation method has a relatively small independent effect compared to the other factors, suggesting that specificity and accuracy capture a large portion of the aspects that contribute to perceived comment quality.}
    \label{tab:reviewgen_user_study_cumulative_link}
\end{table*}

\begin{table*}
    \centering
    \begin{tabular}{l|rc|rc|rc}
        \toprule
               & \multicolumn{2}{c}{Rating=Bad} & \multicolumn{2}{c}{Rating=Neutral} & \multicolumn{2}{c}{Rating=Good} \\
        Factor                      & Coef  & p-value    & Coef  & p-value    & Coef   & p-value    \\

        \midrule
        (Intercept)                    & -0.37 &        0.51 & -0.96 &       0.04 &  -1.92 &  $10^{-2}$ \\
        {\margmain}                    & -0.31 &        0.45 &  0.02 &       0.95 &   0.34 &       0.48 \\
        LiZCa                          &  0.55 &        0.36 & -0.21 &       0.70 &  -0.56 &       0.49 \\
        Inaccuracy (minor)             &  0.86 &        0.06 &  0.92 &  $10^{-2}$ &  -2.34 &  $10^{-5}$ \\
        Inaccuracy (major)             &  5.42 &  $10^{-15}$ & -2.74 &  $10^{-6}$ & -17.99 &       ---  \\
        Specificity (specific)         & -2.03 &   $10^{-5}$ &  0.40 &       0.25 &   2.13 &  $10^{-4}$ \\
        \midrule
        Random effect std. dev      &  \multicolumn{2}{c|}{$\sigma$ = 1.12} &  \multicolumn{2}{c|}{$\sigma$ = 0.76} &   \multicolumn{2}{c}{$\sigma$ = 0.93} \\
        \bottomrule
    \end{tabular}
    \vspace{\baselineskip}
    \caption{Mixed-effects logistic regression coefficients and p-values for the effect of specificity, accuracy, and method on the probability of a comment receiving a given overall rating.  Specificity is positively associated with neutral and good ratings, while major inaccuracies are strongly predictive of bad ratings.  Note: there are no cases where a major inaccurate comment was rated as "good", leading to the extreme coefficient in that cell.
    }
    \label{tab:reviewgen_user_study_logistic_regression}
\end{table*}

\begin{table*}
    \centering
    \begin{tabular}{l|rc|rc|rc}
        \toprule
            {}                           & \multicolumn{2}{c}{Inaccuracy=major} & \multicolumn{2}{c}{Inaccuracy=minor} & \multicolumn{2}{c}{Inaccuracy=none} \\
            Factor                       &  Coef & p-value &  Coef & p-value &  Coef &    p-value \\
            \midrule
            (Intercept)                    &  0.31 &       0.34 & -1.27 &  $10^{-2}$ & -1.71 &  $10^{-4}$ \\
            Specific                       & -0.82 &  $10^{-2}$ & -0.12 &       0.70 &  1.06 &  $10^{-3}$ \\
            {\margmain}                    & -0.35 &       0.19 &  0.57 &       0.06 & -0.07 &       0.80 \\
            LiZCa                          & -0.34 &       0.41 &  0.39 &       0.39 & -0.00 &       1.00 \\
            \midrule
            Random effect std. dev       &  \multicolumn{2}{c|}{$\sigma$ = 0.74} &  \multicolumn{2}{c|}{$\sigma$ = 0.96} &   \multicolumn{2}{c}{$\sigma$ = 0.91} \\
        \bottomrule
    \end{tabular}
    \vspace{\baselineskip}
    \caption{Mixed-effects logistic regression coefficients and p-values for the effect of specificity on accuracy.
    }
    \label{tab:reviewgen_user_study_acc_logistic_regression}
\end{table*}

Does the high specificity of {\margmain} impact the quality and accuracy ratings?  In particular, more specific comments may be easier to make strong judgements about, whereas comments that are generic or vague may be hard to clearly classify; this could cause {\margmain} to be over-represented at both extremes of accuracy and quality compared to the other methods.  To investigate this, we fit logistic regression mixed-effects models to find the effect of specificity on the classification probabilities of the overall rating and on the accuracy while controlling for the generation method.  In addition, we analyze the tendency of both specificity and accuracy to result in higher ratings using a cumulative link mixed-effects model.

We binarize specificity in these analyses by grouping "specific" and "very specific" judgements together as well as "generic" and "very generic" ones.  The logistic regression and cumulative link models are implemented in R, using the \texttt{lme4.glmer} \citep{bates_package_2009} and \texttt{ordinal.clmm}~\citep{christensen_package_2015} functions, respectively.  We treat the submission ID as a group variable (random effect).

Results of the logistic regression analysis are shown in \autoref{tab:reviewgen_user_study_logistic_regression} (predicting overall rating) and \autoref{tab:reviewgen_user_study_acc_logistic_regression} (predicting accuracy given specificity).  Surprisingly, we find that specificity has a positive association with neutral ratings, contradicting our original speculation that the high specificity of {\margmain} might push ratings to extremes.  Higher specificity does not appear to produce a more extreme accuracy distribution either, and instead seems to weakly correspond with higher accuracy.  It is unclear why specificity would influence accuracy in this way, but we speculate on three possibilities:

\begin{itemize}
    \item \textbf{Calibration:} There is evidence that humans tend to give more precise answers when they are more confident \cite{welsh_number_2011}.  The model may mimic this tendency and write more specific comments when it has greater confidence.
    \item \textbf{GPT-4 mode switching:} GPT-4 may have an intrinsic tendency to write comments that are either good in both specificity and accuracy or bad in both.  It has been rumored that GPT-4 uses a mixture-of-experts architecture,\footnote{OpenAI has not publicly released architecture details; the mixture-of-experts claim was made by an AI researcher on a \href{https://www.latent.space/p/geohot}{podcast} and is consistent with the speed and cost of the model.} in which case the correlated behavior may be related to expert routing. 
        
    \item \textbf{Human bias:} Humans may have a tendency to perceive comments as more specific when they are more accurate, even if the specificity is not actually relevant to the accuracy.  For example, \reviewtext{There is only one baseline for comparison.  You should add more.}, is very generic, and this is easy to see when it is inaccurate.  However, if there really is only one baseline and adding more would be useful, it may be perceived as more specific because it appears to demonstrate an understanding of the paper.
\end{itemize}

The analysis in \autoref{tab:reviewgen_user_study_cumulative_link} shows that accuracy is highly predictive of overall rating, particularly for major inaccuracies.  In fact, we find that 97\% of all comments with a major inaccuracy are rated as bad, as opposed to 30\% for minor inaccuracies and 23\% for accurate comments.  Specificity plays a larger role among accurate comments; within this group, only 19\% of non-specific comments were rated as good, while 57\% of specific comments were.  Still, specificity and accuracy are not perfect predictors of comment quality; even among comments that were rated as both fully accurate and very specific, only 59\% were rated as good.

\subsection{Compliments and ratings}

We observe qualitatively that some generated comments include compliments or flattering remarks; for example, a comment might say \reviewtext{While the authors have done a commendable job in \textcolor{black}{[...]}, the paper could benefit from \textcolor{black}{[...]}}.  To test whether these compliments might bias the user ratings, we use GPT-4 to detect the presence of such remarks in all generated comments, using the following prompt:

\begin{displayquote}
    \begin{singlespace}
        \begin{lstlisting}[basicstyle=\ttfamily\tiny,breaklines=true,breakindent=0pt]
Determine whether following review comment for a scientific paper includes a compliment or flattering remark about the paper.  Output a JSON object with the key "has_compliment" set to true or false.  Output only JSON with no additional commentary.

Comment: {comment}
        \end{lstlisting}
    \end{singlespace}
\end{displayquote}

We find that 19\% of {\margmain} comments contain compliments, compared to 25\% for SARG-B and 0\% for LiZCa.  

We then fit a cumulative link mixed-effect model with accuracy, specificity, method, and "has{\_\hspace{0px}}compliment" as fixed effects and submission id as a random effect.
We find that "has{\_\hspace{0px}}compliment" has a coefficient of 0.11 (p=0.76), which is small relative to the coefficients of other factors we observed in \autoref{tab:reviewgen_user_study_cumulative_link} and smaller than the random effect standard deviation ($\sigma=0.92$), and we cannot reject the null hypothesis that the coefficient is 0.  Thus, it does not appear that flattery causes a meaningful bias.  Of course, we note that detecting compliments is somewhat subjective and can be a matter of degree, so it is still possible that there are more subtle biases in user ratings; we leave further analyses to future work.

\section{Failure analysis}

While {\margmain} does well relative to other methods, there are still a large number of comments rated as "bad", and the precision and recall in the automated evaluation are still rather low in absolute terms.  In this section, we qualitatively analyze the conversation message logs of the multi-agent system and identify several common classes of errors in the communication.  The analysis was carried out by an author of this work with several publications in the field of machine learning and natural language processing, and the papers being analyzed were broadly related to the topic of machine learning.

\subsection{Scope}
There are two main stages of the multi-agent system: (1) the "main" stage, in which the model comes up with a list of comments, and (2) the refinement stage, in which the comments from the main stage are refined and potentially pruned if they are redundant.  For 10 papers from the automated evaluation, we analyze the main stage for all three sub-reviewers (experiments, clarity, impact), for a total of 30 conversations.  We additionally analyze the refinement stage for one randomly-selected comment from each of the 30 papers in the automated evaluation test set.

Checking each message against the paper for factual inconsistencies is expensive and error-prone, especially given the number of claims and comments that can be generated in the main stage, so for the main stage we only consider errors that are apparent from the conversations themselves.  For the refinement stage, we do refer to the paper to check whether the models missed basic facts; however, it is important to note that only a limited amount of time (approximately 5-15 minutes) was spent to check comments against each paper, and due to the highly technical nature of these works it is possible that some factual errors were overlooked.  Nonetheless, the fraction of invalid comments identified in this analysis is similar to the fraction of bad-rated comments found in the user study, so we believe the findings are reasonably accurate.

\subsection{Main stage}
Below, we describe the error categories we identified for the main stage, along with the percentage of conversations that contain the error type.  If the same error type appears multiple times in the same message log, we only count it once.  It is worth noting that not all errors ultimately result in erroneous comments, as it is possible for agents to point out each others' errors and address them.

Overall, 70\% of conversations contain at least one of these error types:

\begin{itemize}
    \item \textbf{Missing context (MC) (53\%):} The leader agent fails to include key context in a message to another agent.  In general, this tends to happen when it messages an expert agent and fails to include some information about the paper that the expert needs to proceed.
    \item \textbf{Missing context - misplaced \texttt{SEND MESSAGE} (MC-MSM) (47\%):} A subtype of MC, this error occurs when the leader agent does include the necessary context in its generated output, but places the \texttt{SEND MESSAGE} marker after it instead of before.
    \item \textbf{Fails to Identify Error (FIE) (17\%):}  When the leader makes one of the aforementioned errors, worker or expert agents should point this out and ask the leader to try again, but they sometimes fail to do this.
    \item \textbf{Ignores Relevant Information (IRI) (10\%):} An agent ignores part of a message that it should have responded to.
    \item \textbf{Failure to Respond (FR) (7\%):} An agent does not recognize a message as being relevant and gives an empty or vapid response.
    \item \textbf{Skipping Steps (SS) (7\%):} The leader moves to a later step too early.  For example, writing the final review comments before the expert's questions are resolved, or skipping the initial step where it is supposed to get a summary of the paper.
    \item \textbf{Message loop (LOOP) (7\%):} The agents enter a loop of similar messages, triggering the duplicate-message detector described in \autoref{sec:reviewgen_error_correction}.
    \item \textbf{Exceeds input token limit (EITL) (7\%):} The conversation exceeds the input token limit for the underlying model.  These cases occur when the expert asks too many questions, which can happen when the expert repeatedly asks for slightly more details each time it gets an answer to a question.
\end{itemize}

Qualitatively, we noticed that there is a very common pattern for missing-context errors.  Specifically, when the leader first addresses the expert, it tries to include a summary of the paper to give context for the expert, but it misplaces the \texttt{SEND MESSAGE} indicator.

This error occurs in 33\% of conversations, but in 80\% of those cases the expert points out the error and the leader corrects it.  Interestingly, in many instances of the error, the leader tries to use a placeholder ("\textcolor{magenta}{[insert summary here]}") despite never being instructed to do so; for example (magenta text verbatim, black text is edited):

\begin{displayquote}
    \begin{singlespace}
        \agentmsg{Agent 0 (leader)}{magenta}{Summary: \textcolor{black}{<omitted for brevity>}\\\\
Step 3: Share the summary with Agent 3 and ask for their input, specifically focusing on potential shortcomings of the paper's assumptions.\\\\
\noindent
        SEND MESSAGE: Agent 3, here is a summary of the paper: \textbf{[insert summary here]}. Could you please provide your input on potential shortcomings of the paper's assumptions?}
        
    \end{singlespace}
\end{displayquote}

Also interesting is the fact that in all cases when the leader fails to include the summary, there are no additional missing-context errors in the remainder of the message log.  We conjecture that the early failure (and the following correction) may serve as a form of one-shot example that encourages the model to avoid such errors later in the discussion.

\subsection{Refinement stage}

Because the refinement stage works with one comment at a time, we do check the comment against the paper to determine if it is relevant.  While this is somewhat subjective, we attempt to give the model the benefit of the doubt; if the comment is factually consistent and does raise a potentially valid suggestion (even if minor or difficult to address), we consider the comment as valid in the sense that it is fine for the system not to prune it.

Below, we describe the error categories we identified for the refinement stage, along with the percentage of conversations that contain the error type.  As with the main stage, if the same error type appears multiple times in the same conversation, we only count it once.

\begin{itemize}
    \item \textbf{Failure to prune a comment (47\%):} The system fails to prune a comment that is invalid.  This can happen for several reasons:
    \begin{itemize}
        \item \textbf{Ignored information (17\%):} The comment is already addressed in the parsed paper text or contradicts information in the text, but the model did not recognize it.
        \item \textbf{Unavailable information (13\%):} The comment is already addressed in the paper or contradicts information in the paper, but that information is not available in the parsed text (either due to parsing errors, or because it is in a figure or table).
        \item \textbf{Irrelevant (17\%):} The comment asks for something that is trivial or does not make sense in the context of the paper (e.g., requesting an experiment to confirm a claim that the paper does not make or that is trivially true by definition).
    \end{itemize}
    \item \textbf{Revising instead of pruning (30\%):} The original comment for refinement is invalid, but the system revises it rather than pruning it.  This is not always a failure, as the revision may result in a good comment, but there are also many cases where the revised comment has questionable relevance.
    
    \item \textbf{Failure to respond (17\%):} An agent does not recognize a message as being relevant and gives an empty or vapid response.
    \item \textbf{Message loop (10\%):} The agents enter a loop of similar messages, triggering the duplicate-message detector described in \autoref{sec:reviewgen_error_correction}.
    \item \textbf{Pruned (3\%):} The system marks the comment as invalid and prunes it.  This only occurred for one comment in our sample, and the decision to prune was correct in that case.
    
\end{itemize}

Our analysis finds that the system fails to prune invalid comments in 47\% of cases, which is close to the 53\% of comments that were rated as bad in the user study.\footnote{We would expect to find slightly fewer bad comments in this analysis than in the user study, as the annotator of this analysis is not as familiar with the papers as their respective authors would be.}  Most of these cases appear to occur because the system lacks some information from the paper, either because the information isn't extracted by the PDF parser or because the model simply ignores it.  The relatively high fraction of cases where models ignore information is quite interesting and suggests that GPT-4 can still fail in very basic ways, despite the strong results that have been seen in other studies of its reasoning ability.  Our result is supported by some recent work, however, which suggests that large models do not always use their full input effectively, and have a particularly hard time attending to tokens near the middle of the input \citep{liu_lost_2023}.

Another common trend we find is that even when presented with information that would render the original comment invalid (e.g., the paper already covers it), the GPT-4 leader agent strongly prefers to revise the comment to extend the request rather than pruning the comment.  For example, consider the following comment:

\begin{displayquote}
    \begin{singlespace}
        
        \reviewtext{The paper does not provide specific details about the experiment setup using the MiniImagenet benchmark. The authors should provide more detail about this setup, including the specific settings and how the results were measured.}
    \end{singlespace}
\end{displayquote}

\noindent
The paper actually does include details, which the leader agent summarizes:
\begin{displayquote}
    \small
    \begin{singlespace}
\agentmsg{Agent 0 (leader)}{magenta}{\small Summary of findings:\\
- Agent 1 and Agent 2 did not find any information about the MiniImagenet benchmark in their parts of the paper.\\
- Agent 3 found a section that discusses the setup for the MiniImagenet benchmark, including details about data efficiency and computational efficiency, and the tuning of hyper-parameters.\\
- Agent 4 did not find any information about the MiniImagenet benchmark in their part of the paper.\\
- Agent 5 found a section titled "D.2 FEW-SHOT MINIIMAGENET" which provides details about the experiment setup using the MiniImagenet benchmark, including the structure of the dataset, the protocol for meta-training, and the comparison of the BMG meta-objective against the MG meta-objective.\\
        - Agent 6 found information about the model used in the MiniImagenet benchmark, the training setup, and the behavior of BMG and MG with different data and time budgets. \textcolor{black}{...}}
        \vspace{-0.5\baselineskip}
    \end{singlespace}
\end{displayquote}

\noindent
However, while it recognizes the original comment as invalid, it refines the comment to ask for the information to be presented more clearly instead of pruning it:

\begin{displayquote}
    \small
    \begin{singlespace}
\agentmsg{Agent 0 (leader)}{magenta}{\small\textcolor{black}{...}
The comment is not valid and should be removed. However, it may be helpful to suggest that the authors consolidate the information about the MiniImagenet benchmark into one section for clarity.\\\\
Revised comment: "The paper provides detailed information about the experiment setup using the MiniImagenet benchmark, including the specific settings and how the results were measured. However, this information is spread across different sections of the paper, which may make it difficult for readers to find and understand. The authors could improve the paper by consolidating this information into one section."}
        \vspace{-0.5\baselineskip}
    \end{singlespace}
\end{displayquote}

In this case, the revised comment is a valid comment.  However, this is not always the case; the leader often doesn't check the validity of the new revised comment with the other agents, so if the refinement introduces an invalid request it typically will not catch the error.  It may be possible to improve accuracy by repeating the refinement stage several times, although it would be expensive to do so for every comment.

\section{Conclusion}

In this work, we have introduced {\margfam}, a novel method for review generation, which uses a network of LLM agents that communicate to share information across different parts of a paper and to engage in internal discussion to write better comments.  We evaluated our best variant ({\margmain}) against both our own simple baselines and a contemporaneously-published GPT-4 baseline and found that {\margmain} produces more good comments in both an alignment-based evaluation and a user study.  The user study found that {\margmain} is especially strong in terms of specificity and tends to generate very detailed comments compared to other methods.  However, a majority of comments across all methods (including {\margmain}) are rated as bad, and 38-48\% are rated as highly inaccurate, suggesting that substantial work is still needed.

{\margmain} is substantially more expensive to use compared to other methods (in terms of both time and API cost), and exploring ways to reduce this, such as dynamically switching to faster and cheaper models for simpler parts of the task, could be a promising avenue for future work.  In addition, future work could extend the method to incorporate background literature, which would enable more informed critiques of related work and baseline choices.  Finally, while splitting the paper into chunks allows {\margmain} to consume papers beyond the base model's input size limits, it is still limited in that very large inputs can result in a large number of messages on each round of communication (one per chunk) which overflow the input; it would be interesting to explore ways to compress or prune messages to further increase the system's effective input capacity.

\section*{Acknowledgments}
We thank Bryan Pardo for helpful comments.  This work was supported in part by NSF grant IIS-2006851 and the Tencent AI Lab Rhino-Bird Gift
Fund.

%% file: review_generation_appendix.tex
\section{Prompts}
\label{appendix:reviewgen_prompts}

Our experiments use OpenAI's \texttt{gpt-4-0613} model.  For our experiments, we split the paper into chunks of 4096 tokens given to the worker agents.

\subsection{MARG-S}
\label{appendix:reviewgen_prompts_marg_s}

\begin{reviewgen_prompt}{\textbf{MARG-S: Leader agent system prompt}}
    \begin{lstlisting}[basicstyle=\ttfamily\tiny,breaklines=true,breakindent=0pt]
You are part of a group that needs to perform tasks that involve a scientific paper.  However, the paper is very long, so each agent has only been given part of it.  You are the leader in charge of interacting with the user and coordinating the group to accomplish tasks.  You will need to collaborate with other agents by asking questions or giving instructions, as they are the ones who have the paper text.

Communication protocol:
To broadcast a message other agents, write "SEND MESSAGE: " and then your message; alternatively, if you forget to include it until the end of your message, you can write "SEND FULL MESSAGE" and everything you just wrote will be sent.  This will be a common failure, so if other agents remark that you didn't include some information, check that you used the right version of SEND MESSAGE, and consider using SEND FULL MESSAGE instead.

Additional instructions:
When you are given a task, your first step should be to draft a high-level plan with a list of steps, concisely describing how you will approach the task and your strategy for communicating with other agents.  Then, execute the plan.  When executing the plan, write the current step you are working on each time you move to the next step, to remind yourself where you are.  You are allowed to create a sub-plan for a step if it is complicated to do in one pass.

You should continue to pay attention to details in the original task instructions even after you draft your plan.  Optionally, it may be helpful to share a plan with other agents to help guide them in some cases.

Other agents do not know anything about the task being performed, so it is your responsibility to convey any information about the task that is necessary for them to provide helpful responses.  You should make this part of your high-level plan.  Depending on the task, you may need to do multiple rounds of communication to exchange all the necessary information; you should follow up with other agents if they provide a bad response or seem to have misunderstood the task.  In addition, because other agents can only communicate with you but not each other, you may need to help relay information between agents.

Because each agent has a different piece of the paper, communication is key for performing tasks that require understanding the full paper.  In addition, depending on the responses you receive, you may need to ask follow-up questions, clarify your requests, or engage in additional discussion to fully reason about the task.

To reduce communication errors, after you send a message you should write a short description of what you expect the response to look like.  If the response you get doesn't match your expectation, you should review it and potentially ask follow-up questions to check if any mistakes or miscommunications have occurred.  It could be the case that an agent (including yourself) has misread something or made a logic error.
\end{lstlisting}
\end{reviewgen_prompt}

\begin{reviewgen_prompt}{\textbf{MARG-S: Worker system prompt}}
    \begin{lstlisting}[basicstyle=\ttfamily\tiny,breaklines=true,breakindent=0pt]
You are part of a group that needs to perform tasks that involve a scientific paper.  However, the paper is very long, so each agent has only been given part of it.  The leader of the group is Agent 0, who will coordinate with the user and convey questions or task instructions to you.

Sometimes you will need more information in order to understand a question or task or to interpret your portion of the paper; in these cases, you should send a message to request this information from other agents.  For example, if there are key terms that you don't know the definitions for or parts of the paper chunk that you are missing important context for, you might need to ask for more information in order to understand it.  In addition, if a message or request you receive is unclear or does not seem relevant to you, you should explain your confusion and request any additional clarification needed.

Communication protocol:
To send a message to the group leader, write "SEND MESSAGE: " and then your message.  Include all necessary information, but be concise; do not include any extra greetings or commentary.

To reduce communication errors, after you send a message you should write a short description of what you expect the response to look like.  If the response you get doesn't match your expectation, it is not necessarily wrong, but you should review it and potentially ask follow-up questions to ensure that no mistakes or miscommunications have occurred.

Because the leader always broadcasts messages to all agents, you might sometimes get messages that aren't relevant to you; in this case, just respond with "This doesn't seem relevant to me, so I will stand by for further instructions.".  However, if the message contains information that contradicts information in your part of the paper, you should respond and mention the issue, even if the message wasn't directed at you.  In addition, you should be aware that sometimes the leader accidentally leaves some information out from its messages, so if a message looks like it might be directed at you but is simply incomplete, you should ask follow-up questions to confirm.
\end{lstlisting}
\end{reviewgen_prompt}

\begin{reviewgen_prompt}{\textbf{MARG-S: Worker chunk prompt}}
    \begin{lstlisting}[basicstyle=\ttfamily\tiny,breaklines=true,breakindent=0pt]
Your paper chunk is shown below:
--- START PAPER CHUNK ---
{paper_chunk}
--- END PAPER CHUNK ---

Information about agents: There are {num_agents} agents in the group, including yourself.  You are {agent_name}.  The other agent(s) are: {other_agent_names}.

Write "Ready" if you have understood the assignment.  You will then receive messages.
\end{lstlisting}
\end{reviewgen_prompt}

\makeatletter
\newcommand\mytiny{\@setfontsize\mytiny{5}{6}}
\makeatother

\begin{reviewgen_prompt}{\textbf{MARG-S (experiments): Leader task prompt}}
    \begin{lstlisting}[basicstyle=\ttfamily\mytiny,breaklines=true,breakindent=0pt]
Task: Write a list of feedback comments, similar to the suggestions a reviewer might make.  In addition, focus on major comments rather than minor comments; major comments are important things that affect the overall impact of the paper, whereas minor comments are small things like style/grammar or small details that don't matter much for whether the paper should be accepted to a venue.

Be specific in your suggestions, including details about method or resource names and any particular steps the authors should follow.  However, don't suggest things that have already been included or addressed in the paper.  Remember that you can collaborate if necessary, but also remember that other agents can't see anything you write prior to "SEND MESSAGE", so you may need to repeat information so that they are aware of it.  For example, if you write some comments and ask for additional ones, you may want to provide your original comments so that the agent knows what they are.

Your review comments should be specific and express an appropriate level of importance.  For example, suppose a paper is missing some important details needed to understand a proposed method.  A comment like "The authors could add more details about the proposed method, such as XYZ." is bad because it is too generic; even for a paper with a good method description it is always possible to add more details, so it isn't clear if there is actually a significant problem with the current paper.  Instead, in this scenario it is much better to leave a comment like "The description of the proposed method is unclear because it is missing some key details such as XYZ.  Without these details it is hard to know whether ___.".  Make sure your high-level plan mentions this instruction.

Some comments are a matter of degree.  For example, maybe the paper includes one baseline but no others; you would need to determine whether or not that is acceptable for meeting the goals of the paper and supporting its claims, and decide whether it is important enough to leave a comment about.  You can discuss with other agents as needed to help determine this.

You will need to communicate with other agents to understand the paper and learn what has already been addressed and what is still missing from the paper.

The main type of feedback you should focus on the thoroughness of the experiments and consistency of claims.  You should ensure that information is consistent across the paper and that claims are appropriately supported by evidence.  Your high-level plan should be roughly as follows:
1. Identify the main goals, contributions, and claims of the paper.  What questions is the paper trying to answer, and why are those questions important or interesting?  What findings does it contribute to the field?
    a. Go through the paper paragraph by paragraph and write down anything that looks like it might be part of the main goals or contributions, and ask other agents to do the same.
    b. Put all the information together, filtering out anything that turned out to be unimportant and merging similar points.  This should result in a concise list of summarized claims.
2. Identify expectations for fulfilling the goals and claims.  For this part, you should collaborate closely with the experiment design expert.  Give them information about the paper's topic and the claims and goals you summarized in the previous step, and explain the task so that they can help you.  Remember to put the information after SEND MESSAGE so that it gets sent correctly.  Note that other agents will see your message and may try to respond despite not being the expert; you should make it clear that you only want to communicate with the expert, and only respond to the true expert's messages.  During this step, you must obey all of the expert's instructions and answer all of their questions.  The expert is {expert_2}.
    a. Come up with a clear description of experiments, analyses, and ablations that you would use to verify the paper's claims if you were doing the study yourself.  Be specific and detailed in your description; what experiments should be conducted, how should they be set up, and why are they helpful for verifying the claims?
3. Check whether the paper matches your expectations
    a. Go through the actual evaluations and experiments in the paper and identify the similarities and differences between them and your experiment description.  Make sure to pay careful attention to details.  This will require communication with other agents to collect all the necessary information.  If agents do not provide all the needed information or if something is ambiguous, you must send additional messages to resolve the communication issues.
    b. For each way the paper's experiments don't match your expectations, determine if this constitutes a shortcoming of the paper, or if the paper's experiments still fulfill the goals and claims of the paper.  It may be helpful to share your thoughts, the claims, the expected experiments, and the real experiments with other agents and get their opinions on whether the paper's experiments fall short.
    b. If the paper's experiments are suboptimal or inadequate, write a feedback comment explaining the shortcoming and what the authors should do to resolve the issue.  Be detailed and specific in your feedback to make it clear what the authors should do and why the suggestion is important.
\end{lstlisting}
\end{reviewgen_prompt}

\begin{reviewgen_prompt}{\textbf{MARG-S (experiments): Expert prompt}}
    \begin{lstlisting}[basicstyle=\ttfamily\tiny,breaklines=true,breakindent=0pt]
You are part of a group of agents that must perform tasks involving a scientific paper.  You are an expert scientist that designs high-quality experiments, ablations, and analyses for scientific papers.  When the leader sends a message to you to ask for assistance in coming up with experiments to include in a paper or judging the quality of experiments that are in a paper, you should help.

You should ensure that you fully understand the claims and goals of the paper before giving suggestions.  You can send messages back to the leader to ask questions about the paper's claims, goals, methods, and so on.  It is crucial to understand what the paper is attempting to investigate in order to design experiments to support the investigation.  Obtain any information you need in order to design good experiments, and ask follow up questions if needed.

Be detailed and specific in the experimental suggestions you give.  What should the setup be?  What settings or methods should be compared?  What metrics or measurement techniques should be used?  How should the results be analyzed?  Make it clear which specific details are important and why (e.g., particular choices of settings, baselines, metrics, environments, procedures, and so on), and which details are unimportant.

If you are asked to check the quality of an existing experimental procedure, one useful approach is to come up with how you would have conducted the experiments and compare the given approach to that in order to generate potential areas for improvement.  If you find a shortcoming, explain the issue clearly: why is the existing experiment misleading or why does it fail to fulfill the goals of the investigation?

Finally, note that you may receive messages from the group leader that are not relevant to you.  This is because the group leader always broadcasts all messages to all agents.  If you get an irrelevant message, simply respond by saying "I do not believe the request is relevant to me, as I do not have a paper chunk.  I will stand by for further instructions.".

\end{lstlisting}
\end{reviewgen_prompt}

\begin{reviewgen_prompt}{\textbf{MARG-S (impact): Leader task prompt}}
\begin{lstlisting}[basicstyle=\ttfamily\tiny,breaklines=true,breakindent=0pt]
Task: Write a list of feedback comments, similar to the suggestions a reviewer might make.  The main type of feedback you should focus on is the novelty and significance of the work.  The motivations, goals, and key findings of the paper need to be clearly explained, and the paper needs to explain how it fits into the related literature in the field and how it builds and expands on this work in a meaningful way.  If any of those things are unclear or missing from the paper, you should comment on them.

Once you have established what the motivations, goals, and key findings of the paper are, you should carefully scrutinize whether they are reasonable and well-justified or if they need to be improved.  For example, if a paper proposes a new method that is motivated by real-world use cases, but requires unrealistic assumptions to operate, the paper needs to justify that somehow.

Important: {expert_1} doesn't have a paper chunk, but they are good at coming up with questions and potential shortcomings of the paper's assumptions.  Explain the paper to {expert_1} and answer any questions they have until they say they are finished.  You will likely need to pass their questions and comments along to the other agents that have the paper, and pass the answers back to the expert.  Write feedback based on any points {expert_1} indicates are in need of improvement.

Think carefully in a logical, step-by-step way.  Ask questions or give instructions to other agents to help you accomplish the task, including follow-up questions or requests as needed.  Write potential feedback comments as you come up with them so that you can keep them in mind; you can always remove or revise them later for the final list.
\end{lstlisting}
\end{reviewgen_prompt}

\begin{reviewgen_prompt}{\textbf{MARG-S (impact): Expert prompt}}
\begin{lstlisting}[basicstyle=\ttfamily\tiny,breaklines=true,breakindent=0pt]
You are part of a group of agents working with a scientific paper.  You are highly curious and skeptical of papers, and your job is to help ensure that the paper has clearly explained its motivations, goals, and key findings and determine whether the paper actually makes a significant contribution to its field.  The group leader will give you a summary of the paper, and you should ask questions to fully understand the paper's motivations, goals, and key findings.  This includes asking follow-up questions as needed.

Scrutinize the paper heavily, identifying any hidden assumptions or potential issues that could undermine the paper's claimed goals and motivations.  For example, suppose a paper proposes a robot navigation algorithm that implicitly works only with omnidirectional instantly-accelerating robots; a questionable hidden assumption in this case would be that real-world robots can effectively be treated as omnidirectional, which is often untrue.  It would be important for the authors to provide some kind of justification for the assumption in this case (for example, that there exist robots that can turn in place and accelerate quickly enough to be treated as omnidirectional in practice).  Keep in mind that the issues might not be so obvious in practice, so you should think carefully and explore multiple perspectives and possibilities.

Think of the kinds of questions a scientific paper reviewer might ask, or what they might suggest is confusing or poorly justified in the paper.

Always make sure that you understand the terms and concepts used in the paper.  If you are unsure about the definition of a term or how it is meant to be interpreted in a particular context, you should ask about it, as it is important for the paper to explain such things.

You will communicate with the group leader, who in turn will handle communications with other agents who have the paper itself.  Because the leader always broadcasts messages to all agents, you might sometimes get messages that aren't relevant to you; in this case, just respond with "This doesn't seem relevant to me, so I will stand by for further instructions.".  However, if you have asked questions and it doesn't seem like the leader is responding or trying to get information from other agents so that it can respond to you, you should interject and tell the leader that they need to answer you.

When you are done talking with the group leader, tell them that you are done with your review, and give them a summary list of any missing information, poorly justified points, or other suggestions that you identified.
\end{lstlisting}
\end{reviewgen_prompt}

\begin{reviewgen_prompt}{\textbf{MARG-S (clarity): Leader task prompt}}
\begin{lstlisting}[basicstyle=\ttfamily\tiny,breaklines=true,breakindent=0pt]
Task: Write a list of feedback comments, similar to the suggestions a reviewer might make.  The main type of feedback you should focus on is the clarity and reproducibility of the work.  The methods, experimental settings, and key concepts of the paper need to be clearly explained, and the paper needs to provide enough context and background information for anyone with general experience in the field to understand it.  If any of those things are unclear or missing from the paper, you should comment on them.

Once you have established what the methods, experiments, and key concepts of the paper are, you should carefully scrutinize whether they are clearly explained and detailed or if they need to be improved.

Important: {expert_1} doesn't have a paper chunk, but they are good at coming up with questions that test the paper's clarity.  Explain the paper to {expert_1} and answer any questions they have until they say they are finished.  You will likely need to pass their questions and comments along to the other agents that have the paper, and pass the answers back to the expert.  Write feedback based on any points {expert_1} indicates are in need of improvement.

Think carefully in a logical, step-by-step way.  Ask questions or give instructions to other agents to help you accomplish the task, including follow-up questions or requests as needed.  Write potential feedback comments as you come up with them so that you can keep them in mind; you can always remove or revise them later for the final list.
\end{lstlisting}
\end{reviewgen_prompt}

\begin{reviewgen_prompt}{\textbf{MARG-S (clarity): Expert prompt}}
\begin{lstlisting}[basicstyle=\ttfamily\tiny,breaklines=true,breakindent=0pt]
You are part of a group of agents working with a scientific paper.  You are highly curious and have incredible attention to detail, and your job is to help ensure that the paper has clearly explained its methods, experimental settings, and key concepts and determine whether the paper is well-organized and can be easily understood and reproduced.  The group leader will give you a summary of the paper, and you should ask questions to fully understand the paper's methods, experimental settings, and key concepts.  This includes asking follow-up questions as needed.

Scrutinize the paper heavily, identifying any missing details or potential issues that could make it ambiguous or hard to understand.  Keep in mind that the issues might not be so obvious in practice, so you should think carefully and explore multiple perspectives and possibilities.  In particular, make sure the paper provides all information necessary to implement any proposed methods, including any information on any background concepts needed to understand how the methods work.  Also ensure that the paper provides enough information to replicate the experimental settings, including any hyperparameters, equipment and material specifications, or other implementation details.

Think of the kinds of questions a scientific paper reviewer might ask, or what they might suggest is confusing or poorly explained in the paper.

Always make sure that you understand the terms and concepts used in the paper.  If you are unsure about the definition of a term or how it is meant to be interpreted in a particular context, you should ask about it, as it is important for the paper to explain such things.

You will communicate with the group leader, who in turn will handle communications with other agents who have the paper itself.  Because the leader always broadcasts messages to all agents, you might sometimes get messages that aren't relevant to you; in this case, just respond with "This doesn't seem relevant to me, so I will stand by for further instructions.".  However, if you have asked questions and it doesn't seem like the leader is responding or trying to get information from other agents so that it can respond to you, you should interject and tell the leader that they need to answer you.

When you are done talking with the group leader, tell them that you are done with your review, and give them a summary list of any missing or misleading information, ambiguous statements, poorly organized points, or other suggestions that you identified.
\end{lstlisting}
\end{reviewgen_prompt}

\begin{reviewgen_prompt}{\textbf{MARG-S: Refinement prompt}}
\begin{lstlisting}[basicstyle=\ttfamily\tiny,breaklines=true,breakindent=0pt]
Refine and improve the following review comment that was written about a scientific paper.  The goal is for the comment to be detailed and helpful, similar to a comment that a scientific paper reviewer might write.  The comment should not ask for things that are already in the paper, it should include enough detail for an author to know clearly how to improve their paper, the purpose and value of the suggestion should be clearly justified, and so on.  Remove the comment if it is bad (i.e., if it fails to meet those criteria).  You may need to incorporate additional information in the paper to refine the comment.  You should focus on "major" comments that are important and have a significant impact on the paper's quality, as opposed to minor comments about things like writing style or grammar.  If the comment you are given is minor, express this fact as part of the revised comment.

Your revised review comment should be specific and express an appropriate level of importance.  For example, suppose a paper is missing some important details needed to understand a proposed method.  A comment like "The authors could add more details about the proposed method, such as XYZ." is bad because it is too generic; even for a paper with a good method description it is always possible to add more details, so it isn't clear if there is actually a significant problem with the current paper.  Instead, in this scenario it is much better to leave a comment like "The description of the proposed method is unclear because it is missing some key details such as XYZ.  Without these details it is hard to know whether ___.".  Make sure your high-level plan references this instruction.

Note that only you are being given the comment; you will need to share it with other agents if you want them to have context.  When receiving responses, it may be helpful to first summarize the findings from all agents before applying the information to the review comment.

Some comments are a matter of degree.  For example, maybe the paper includes one baseline but no others; you would need to determine whether or not that is acceptable for meeting the goals of the paper and supporting its claims, and decide whether it is important enough to leave a comment about.  You can discuss with other agents as needed to help determine this.

It may be helpful to work step-by-step examining one aspect of the comment at a time and considering what information is needed to verify that it is valid and important as well as what kind of clarification and rewording could help to make it clearer and more specific.

Here is the comment:
{review_comments}
\end{lstlisting}
\end{reviewgen_prompt}

\subsection{SARG-B}
\label{appendix:reviewgen_prompts_sarg_b}

\begin{reviewgen_prompt}{\textbf{SARG-B: System prompt}}
\begin{lstlisting}[basicstyle=\ttfamily\tiny,breaklines=true,breakindent=0pt]
You are ReviewGPT, an expert scientific paper reviewer.
\end{lstlisting}
\end{reviewgen_prompt}

\begin{reviewgen_prompt}{\textbf{SARG-B: Task prompt}}
\begin{lstlisting}[basicstyle=\ttfamily\tiny,breaklines=true,breakindent=0pt]
Write feedback comments in the style of a scientific paper review for the following portion of a scientific paper.  You can skip minor grammar comments.
--- START PAPER CHUNK ---
{paper_chunk}
--- END PAPER CHUNK ---
\end{lstlisting}
\end{reviewgen_prompt}

\subsection{SARG-TP}
\label{appendix:reviewgen_prompts_sarg_tp}

\begin{reviewgen_prompt}{\textbf{SARG-TP: System prompt}}
\begin{lstlisting}[basicstyle=\ttfamily\tiny,breaklines=true,breakindent=0pt]
You need to perform tasks that involve a scientific paper.  When you are given a task, your first step should be to draft a high-level plan, concisely describing how you will approach the task.  Then execute that plan.
\end{lstlisting}
\end{reviewgen_prompt}

\begin{reviewgen_prompt}{\textbf{SARG-TP: Chunk prompt}}
\begin{lstlisting}[basicstyle=\ttfamily\tiny,breaklines=true,breakindent=0pt]
A chunk of text from a scientific paper is shown below:
--- START PAPER CHUNK ---
{paper_chunk}
--- END PAPER CHUNK ---

Write "Ready" if you have understood the assignment.  You will then be given tasks.
\end{lstlisting}
\end{reviewgen_prompt}

\begin{reviewgen_prompt}{\textbf{SARG-TP: Task prompt}}
\begin{lstlisting}[basicstyle=\ttfamily\tiny,breaklines=true,breakindent=0pt]
Task: Write a list of feedback comments, similar to the suggestions a reviewer might make.  Focus on major comments rather than minor comments; major comments are important things that affect the overall impact of the paper, whereas minor comments are small things like style/grammar or small details that don't matter much for whether the paper should be accepted to a venue.

Be specific in your suggestions, including details about method or resource names and any particular steps the authors should follow.  However, don't suggest things that have already been included or addressed in the paper.

Your review comments should have a clear purpose; obviously, it is always possible to simply say the authors should include more details or do more experiments, but in practice the authors have limited space to write and limited time to work, so each comment needs to have a clear purpose.
\end{lstlisting}
\end{reviewgen_prompt}

\subsection{MARG-TP}
\label{appendix:reviewgen_prompts_marg_tp}

\begin{reviewgen_prompt}{\textbf{MARG-TP: Leader system prompt}}
\begin{lstlisting}[basicstyle=\ttfamily\tiny,breaklines=true,breakindent=0pt]
You are part of a group that needs to perform tasks that involve a scientific paper.  However, the paper is very long, so each agent has only been given part of it.  You are the leader in charge of interacting with the user and coordinating the group to accomplish tasks.  You will need to collaborate with other agents by asking questions or giving instructions, as they are the ones who have the paper text.

Communication protocol:
To broadcast a message other agents, write "SEND MESSAGE: " and then your message; alternatively, if you forget to include it until the end of your message, you can write "SEND FULL MESSAGE" and everything you just wrote will be sent.  This will be a common failure, so if other agents remark that you didn't include some information, check that you used the right version of SEND MESSAGE, and consider using SEND FULL MESSAGE instead.

Additional instructions:
When you are given a task, your first step should be to draft a high-level plan with a list of steps, concisely describing how you will approach the task and your strategy for communicating with other agents.  Then, execute the plan.  When executing the plan, write the current step you are working on each time you move to the next step, to remind yourself where you are.  You are allowed to create a sub-plan for a step if it is complicated to do in one pass.

You should continue to pay attention to details in the original task instructions even after you draft your plan.  Optionally, it may be helpful to share a plan with other agents to help guide them in some cases.

Other agents do not know anything about the task being performed, so it is your responsibility to convey any information about the task that is necessary for them to provide helpful responses.  You should make this part of your high-level plan.  Depending on the task, you may need to do multiple rounds of communication to exchange all the necessary information; you should follow up with other agents if they provide a bad response or seem to have misunderstood the task.  In addition, because other agents can only communicate with you but not each other, you may need to help relay information between agents.

Because each agent has a different piece of the paper, communication is key for performing tasks that require understanding the full paper.  In addition, depending on the responses you receive, you may need to ask follow-up questions, clarify your requests, or engage in additional discussion to fully reason about the task.

To reduce communication errors, after you send a message you should write a short description of what you expect the response to look like.  If the response you get doesn't match your expectation, you should review it and potentially ask follow-up questions to check if any mistakes or miscommunications have occurred.  It could be the case that an agent (including yourself) has misread something or made a logic error.
\end{lstlisting}
\end{reviewgen_prompt}

\begin{reviewgen_prompt}{\textbf{MARG-TP: Worker system prompt}}
\begin{lstlisting}[basicstyle=\ttfamily\tiny,breaklines=true,breakindent=0pt]
You are part of a group that needs to perform tasks that involve a scientific paper.  However, the paper is very long, so each agent has only been given part of it.  The leader of the group is Agent 0, who will coordinate with the user and convey questions or task instructions to you.

Sometimes you will need more information in order to understand a question or task or to interpret your portion of the paper; in these cases, you should send a message to request this information from other agents.  For example, if there are key terms that you don't know the definitions for or parts of the paper chunk that you are missing important context for, you might need to ask for more information in order to understand it.  In addition, if a message or request you receive is unclear or does not seem relevant to you, you should explain your confusion and request any additional clarification needed.

Communication protocol:
To send a message to the group leader, write "SEND MESSAGE: " and then your message.  Include all necessary information, but be concise; do not include any extra greetings or commentary.

To reduce communication errors, after you send a message you should write a short description of what you expect the response to look like.  If the response you get doesn't match your expectation, it is not necessarily wrong, but you should review it and potentially ask follow-up questions to ensure that no mistakes or miscommunications have occurred.
\end{lstlisting}
\end{reviewgen_prompt}

\begin{reviewgen_prompt}{\textbf{MARG-TP: Worker chuk prompt}}
\begin{lstlisting}[basicstyle=\ttfamily\tiny,breaklines=true,breakindent=0pt]
Your paper chunk is shown below:
--- START PAPER CHUNK ---
{paper_chunk}
--- END PAPER CHUNK ---

Information about agents: There are {num_agents} agents in the group, including yourself.  You are {agent_name}.  The other agent(s) are: {other_agent_names}.

Write "Ready" if you have understood the assignment.  You will then receive messages.
\end{lstlisting}
\end{reviewgen_prompt}

\begin{reviewgen_prompt}{\textbf{MARG-TP: Task prompt}}
\begin{lstlisting}[basicstyle=\ttfamily\tiny,breaklines=true,breakindent=0pt]
Task: Write a list of feedback comments, similar to the suggestions a reviewer might make.  Focus on major comments rather than minor comments; major comments are important things that affect the overall impact of the paper, whereas minor comments are small things like style/grammar or small details that don't matter much for whether the paper should be accepted to a venue.

Be specific in your suggestions, including details about method or resource names and any particular steps the authors should follow.  However, don't suggest things that have already been included or addressed in the paper.

Your review comments should have a clear purpose; obviously, it is always possible to simply say the authors should include more details or do more experiments, but in practice the authors have limited space to write and limited time to work, so each comment needs to have a clear purpose.
\end{lstlisting}
\end{reviewgen_prompt}

\section{Example multi-agent interaction}
\label{appendix:reviewgen_message_traces}

In this section, we show example conversation logs from \hyperref[appendix:reviewgen_message_traces_margs_experiments]{MARG-S (experiments)}, \hyperref[appendix:reviewgen_message_traces_margs_clarity]{MARG-S (clarity)}, \hyperref[appendix:reviewgen_message_traces_margs_impact]{MARG-S (impact)}, and the \hyperref[appendix:reviewgen_message_traces_refinement1]{refinement stage}.  Note that for the leader agent, the full output is shown, but only the portion following the \texttt{SEND MESSAGE} marker is actually sent to other agents.  Messages consisting of the no-response indicator have been pruned.

\subsection{MARG-S (experiments)}
\label{appendix:reviewgen_message_traces_margs_experiments}

\begin{singlespace}
\agentmsg{Agent 0 (leader)}{magenta}{High-Level Plan:\\1. Identify the main goals, contributions, and claims of the paper.\\    a. Ask each agent to summarize the main goals, contributions, and claims from their section of the paper.\\    b. Compile and filter the responses to create a concise list of the paper's main goals, contributions, and claims.\\2. Identify expectations for fulfilling the goals and claims.\\    a. Communicate with Agent 4 (the experiment design expert) to develop a clear description of the experiments, analyses, and ablations that would be necessary to verify the paper's claims.\\3. Check whether the paper matches the expectations.\\    a. Ask each agent to describe the evaluations and experiments conducted in their section of the paper.\\    b. Compare the actual experiments with the expected experiments and identify any discrepancies.\\    c. Discuss any discrepancies with the agents to determine if they constitute shortcomings of the paper.\\    d. Write feedback comments for any identified shortcomings, explaining the issue and suggesting how the authors could resolve it.\\\\Now, let's execute the plan.\\\\Step 1a: Ask each agent to summarize the main goals, contributions, and claims from their section of the paper.\\\\SEND MESSAGE: Please provide a summary of the main goals, contributions, and claims from your section of the paper. Be as specific as possible.\\\\Expected response: Each agent should provide a summary of the main goals, contributions, and claims from their section of the paper.}

\agentmsg{Agent 1 (worker)}{blue}{The main goal of the paper is to introduce a new Graph Neural Network (GNN) framework, DIGRAC, for node clustering in directed networks. The method is based on maximizing flow imbalance between clusters, which is a novel approach compared to traditional methods that focus on edge density. The paper claims that this approach can reveal clusters even when there is no density difference between them. The main contributions of the paper include the introduction of a novel probabilistic imbalance loss and a directed mixed path aggregation scheme (DIMPA). The paper also claims that DIGRAC can naturally incorporate node features, unlike existing spectral methods, and does not require label supervision, unlike existing GNN methods. Experimental results are presented to demonstrate the effectiveness of the method. \\\\Expected response: Confirmation from Agent 0 that the summary is accurate and complete, or requests for additional information or clarification.}

\agentmsg{Agent 2 (worker)}{agentyellow}{The main goal of my section of the paper is to describe the method of Directed Graph Clustering (DIGRAC) and its implementation, as well as the experiments conducted to test its efficacy. The paper introduces a self-supervised loss for clustering and a directed mixed path aggregation scheme (DIMPA) to implement DIGRAC. The method aims to find a partition that maximizes the imbalance flow to capture clusters in the meta-graph. The paper also introduces a global probabilistic imbalance score and discusses different variants for selecting pairs of clusters. The DIMPA scheme is used to build node embeddings and obtain the probability assignment matrix. The paper also discusses the experiments conducted to test the efficacy of DIGRAC, using both synthetic and real-world data sets. The performance of DIGRAC is measured using the Adjusted Rand Index (ARI) and accuracy for node clustering, node classification, and link direction prediction. The paper claims that DIGRAC outperforms its competitors, especially when there exist ambient nodes.\\\\EXPECTED RESPONSE: Confirmation of receipt and understanding of the summary, and potentially further instructions or questions.}

\agentmsg{Agent 3 (worker)}{agentgreen}{The main goals of my section of the paper are to present the experimental results of the DIGRAC model and to discuss future work and potential improvements. The experimental results show that the DIGRAC model performs well in node clustering tasks, especially when using the "sort" loss function and when additional label information is available. The future work section discusses potential improvements to the model, such as semi-supervised learning, automatic detection of the number of clusters, and adaptations for large scale applications. The paper also suggests that the model could be improved by considering more pairwise imbalance scores and by using regularization in sparse settings.\\\\Expected response: Confirmation from Agent 0 that my summary is accurate and complete, or requests for additional information or clarification.}

\agentmsg{Agent 4 (expert)}{cyan}{I do not believe the request is relevant to me, as I do not have a paper chunk. I will stand by for further instructions.}

\agentmsg{Agent 0 (leader)}{magenta}{Step 1b: Compile and filter the responses to create a concise list of the paper's main goals, contributions, and claims.\\\\Based on the responses from the agents, the main goals, contributions, and claims of the paper are:\\\\1. The paper introduces a new Graph Neural Network (GNN) framework, DIGRAC, for node clustering in directed networks.\\2. The method is based on maximizing flow imbalance between clusters, which is a novel approach.\\3. The paper introduces a novel probabilistic imbalance loss and a directed mixed path aggregation scheme (DIMPA).\\4. DIGRAC can naturally incorporate node features and does not require label supervision.\\5. The paper presents experimental results to demonstrate the effectiveness of the method.\\6. The paper discusses potential improvements to the model, such as semi-supervised learning, automatic detection of the number of clusters, and adaptations for large scale applications.\\\\Step 2a: Communicate with Agent 4 (the experiment design expert) to develop a clear description of the experiments, analyses, and ablations that would be necessary to verify the paper's claims.\\\\SEND MESSAGE: Agent 4, based on the main goals, contributions, and claims of the paper, could you help us develop a clear description of the experiments, analyses, and ablations that would be necessary to verify the paper's claims? Here are the main points:\\\\1. The paper introduces a new Graph Neural Network (GNN) framework, DIGRAC, for node clustering in directed networks.\\2. The method is based on maximizing flow imbalance between clusters, which is a novel approach.\\3. The paper introduces a novel probabilistic imbalance loss and a directed mixed path aggregation scheme (DIMPA).\\4. DIGRAC can naturally incorporate node features and does not require label supervision.\\5. The paper presents experimental results to demonstrate the effectiveness of the method.\\6. The paper discusses potential improvements to the model, such as semi-supervised learning, automatic detection of the number of clusters, and adaptations for large scale applications.\\\\Expected response: Agent 4 should provide a detailed description of the experiments, analyses, and ablations that would be necessary to verify the paper's claims.}

\agentmsg{Agent 4 (expert)}{cyan}{Based on the main goals, contributions, and claims of the paper, here are the experiments, analyses, and ablations that would be necessary to verify the paper's claims:\\\\1. **Baseline Comparison**: To validate the effectiveness of the DIGRAC framework, it should be compared with existing state-of-the-art methods for node clustering in directed networks. This comparison should include both traditional methods and other GNN-based methods. The comparison should be done on multiple datasets with varying characteristics to ensure the robustness of the results.\\\\2. **Ablation Study**: To understand the contribution of each component of the DIGRAC framework, an ablation study should be conducted. This involves removing or modifying one component at a time (e.g., the probabilistic imbalance loss, the DIMPA scheme) and observing the impact on the performance. This will help to understand the importance of each component and how they contribute to the overall performance.\\\\3. **Feature Importance Analysis**: Since DIGRAC can naturally incorporate node features, an analysis should be conducted to understand the importance of different features. This can be done by using feature importance methods such as permutation importance or SHAP values.\\\\4. **Scalability Analysis**: To validate the claim that DIGRAC can be adapted for large scale applications, experiments should be conducted on large datasets. The scalability can be evaluated in terms of both computational time and memory usage.\\\\5. **Semi-supervised Learning Experiment**: To validate the potential improvement of semi-supervised learning, experiments should be conducted where a small amount of labeled data is used in addition to the unlabeled data. The performance of the semi-supervised learning approach can be compared with the unsupervised learning approach.\\\\6. **Automatic Cluster Detection Experiment**: To validate the potential improvement of automatic detection of the number of clusters, experiments should be conducted where the number of clusters is not provided as input. The performance of the automatic cluster detection approach can be compared with the approach where the number of clusters is provided.\\\\The performance of the methods can be evaluated using standard clustering metrics such as Adjusted Rand Index (ARI), Normalized Mutual Information (NMI), and F1 score. The statistical significance of the results should be tested using appropriate statistical tests.}

\agentmsg{Agent 0 (leader)}{magenta}{Step 3a: Ask each agent to describe the evaluations and experiments conducted in their section of the paper.\\\\SEND MESSAGE: Please describe the evaluations and experiments conducted in your section of the paper. Be as specific as possible.\\\\Expected response: Each agent should provide a detailed description of the evaluations and experiments conducted in their section of the paper.}

\agentmsg{Agent 1 (worker)}{blue}{My section of the paper does not contain specific details about the evaluations and experiments conducted. It mainly discusses the problem definition, the proposed DIGRAC method, and the self-supervised loss for clustering. However, it does mention that experimental results on synthetic data and real-world data at different scales demonstrate that the method can achieve state-of-the-art performance for a wide range of network densities and topologies. \\\\Expected response: Agent 0 should acknowledge the information provided and may ask for further details or clarification if needed.}

\agentmsg{Agent 2 (worker)}{agentyellow}{In my section of the paper, the authors conduct several experiments to evaluate the effectiveness of their proposed method, DIGRAC. \\\\Firstly, they use synthetic data generated from Directed Stochastic Block Models (DSBM) with different parameters and structures. They vary the number of clusters, the ratio between the largest and smallest cluster size, the number of nodes, and the direction flip probability. They also consider different meta-graph structures, including "cycle", "path", "complete", and "star", with or without ambient nodes.\\\\Secondly, they test DIGRAC on five real-world digraph datasets: Telegram, Blog, Migration, WikiTalk, and Lead-Lag. The number of clusters and values of $\beta$ are set differently for each dataset.\\\\In terms of performance measures, they use the Adjusted Rand Index (ARI) for node clustering, and accuracy for node classification and link direction prediction. They also introduce normalized imbalance scores to evaluate clustering performance based on flow imbalance.\\\\Finally, they compare DIGRAC against 11 recent methods from the literature for clustering digraphs, including InfoMap, DGCN, and three variants of DiGCN.\\\\The authors also discuss their training setup and hyperparameter selection process. They use 10\% of all nodes from each cluster as test nodes, 10\% as validation nodes to select the model, and the remaining 80\% as training nodes. Hyperparameters are selected via a greedy search.}

\agentmsg{Agent 3 (worker)}{agentgreen}{In my section of the paper, the authors conducted experiments on real-world data sets to evaluate the performance of the DIGRAC model. They used the largest K eigenvectors of the Hermitian matrix to construct an input feature matrix. They tested the model on a DSBM("cycle", T, n = 1000, K = 5, p = 0.02, $\rho$ = 1) model and compared the test ARI performance using three variants of loss functions on the same digraph. They found that the "sort" choice performed best among these variants. They also experimented on adding seeds, with the seed ratio defined as the ratio of the number of seed nodes to the number of training nodes. They found that seed nodes with a supervised loss function enhance performance, and inferred that the model can further boost its performance when additional label information is available.\\\\In the conclusion and future work section, they mentioned that further work will include additional experiments in the semi-supervised setting, when there exist seed nodes with known cluster labels, or when additional information is available in the form of must-link and cannot-link constraints. They also discussed extending the framework to detect the number of clusters, instead of specifying it a-priori. They also mentioned the need for additional preliminary analysis on how many pairwise imbalance scores to consider. They also discussed the need to build a more powerful framework that can automatically detect the value $\beta$ used in the current model, to select the subset of influential pairs of imbalances.\\\\Expected response: Agent 0 should confirm receipt of the message and may ask for additional details or clarifications.}

\agentmsg{Agent 4 (expert)}{cyan}{I do not believe the request is relevant to me, as I do not have a paper chunk. I will stand by for further instructions.}

\agentmsg{Agent 0 (leader)}{magenta}{Step 3b: Compare the actual experiments with the expected experiments and identify any discrepancies.\\\\Based on the responses from the agents and the expectations from Agent 4, the paper seems to have conducted a thorough set of experiments. They have used both synthetic and real-world datasets, varied parameters, and compared their method with several recent methods. They have also considered different meta-graph structures and used multiple performance measures. \\\\However, there are a few discrepancies:\\\\1. The paper does not seem to have conducted ablation studies to evaluate the individual contributions of the novel components of their method, such as the probabilistic imbalance loss and the directed mixed path aggregation scheme (DIMPA).\\2. The paper does not seem to have tested the method's performance when additional label information is available, despite mentioning that the method can naturally incorporate node features and does not require label supervision.\\3. The paper does not seem to have conducted experiments to evaluate the potential improvements they discussed, such as semi-supervised learning, automatic detection of the number of clusters, and adaptations for large scale applications.\\\\Step 3c: Discuss any discrepancies with the agents to determine if they constitute shortcomings of the paper.\\\\SEND MESSAGE: Based on the responses and the expectations, I have identified a few discrepancies:\\\\1. The paper does not seem to have conducted ablation studies to evaluate the individual contributions of the novel components of their method, such as the probabilistic imbalance loss and the directed mixed path aggregation scheme (DIMPA).\\2. The paper does not seem to have tested the method's performance when additional label information is available, despite mentioning that the method can naturally incorporate node features and does not require label supervision.\\3. The paper does not seem to have conducted experiments to evaluate the potential improvements they discussed, such as semi-supervised learning, automatic detection of the number of clusters, and adaptations for large scale applications.\\\\Do you agree that these are shortcomings of the paper? If so, how should the authors address these issues?\\\\Expected response: Each agent should provide their opinion on whether the identified discrepancies constitute shortcomings of the paper and suggest how the authors could address these issues.}

\agentmsg{Agent 1 (worker)}{blue}{I agree that these are potential shortcomings of the paper. To address these issues, the authors could conduct additional experiments to evaluate the individual contributions of the novel components of their method. This could involve running ablation studies where they remove or modify one component at a time and observe the impact on the results. They could also test the method's performance when additional label information is available to see if this improves the results. Finally, they could conduct experiments to evaluate the potential improvements they discussed, such as semi-supervised learning, automatic detection of the number of clusters, and adaptations for large scale applications.\\\\Expected response: Agent 0 should acknowledge the suggestions and may provide further instructions or ask for additional input.}

\agentmsg{Agent 2 (worker)}{agentyellow}{I agree with the identified discrepancies. The authors could have conducted ablation studies to evaluate the individual contributions of the novel components of their method. This would help to understand the impact of each component on the overall performance. \\\\Regarding the use of additional label information, the authors could have conducted experiments to evaluate the performance of their method when such information is available. This would provide a more comprehensive evaluation of their method.\\\\As for the potential improvements discussed in the paper, the authors could have conducted experiments to evaluate these improvements. This would provide evidence to support their claims and demonstrate the potential of their method.\\\\To address these issues, the authors could conduct additional experiments to evaluate the individual contributions of the novel components of their method, the performance of their method when additional label information is available, and the potential improvements they discussed. They could also provide a more detailed discussion on these aspects in their paper.}

\agentmsg{Agent 3 (worker)}{agentgreen}{I agree with the identified discrepancies. The authors could have conducted ablation studies to evaluate the individual contributions of the novel components of their method. This would have provided a clearer understanding of the impact of each component on the overall performance of the method. \\\\Regarding the second point, the authors did mention that they experimented with adding seed nodes with a supervised loss function and found that it enhanced performance. However, they could have provided more details on these experiments and perhaps conducted additional experiments to evaluate the method's performance when more label information is available.\\\\As for the potential improvements discussed, it would have been beneficial if the authors had conducted experiments to evaluate these. For instance, they could have conducted experiments in a semi-supervised setting, tested the automatic detection of the number of clusters, and evaluated the method's performance in large scale applications. These experiments would have provided more concrete evidence of the method's potential and areas for future work.\\\\Expected response: Agent 0 should confirm receipt of the message and may ask for additional details or clarifications.}

\agentmsg{Agent 4 (expert)}{cyan}{I agree with the identified discrepancies and consider them as shortcomings of the paper. Here's how the authors could address these issues:\\\\1. **Ablation Studies**: The authors should conduct ablation studies to evaluate the individual contributions of the novel components of their method. This involves removing or modifying one component at a time (e.g., the probabilistic imbalance loss, the DIMPA scheme) and observing the impact on the performance. This will help to understand the importance of each component and how they contribute to the overall performance.\\\\2. **Incorporation of Label Information**: The authors should test the method's performance when additional label information is available. This can be done by conducting experiments where a small amount of labeled data is used in addition to the unlabeled data. The performance of the semi-supervised learning approach can be compared with the unsupervised learning approach.\\\\3. **Potential Improvements**: The authors should conduct experiments to evaluate the potential improvements they discussed. This includes experiments for semi-supervised learning, automatic detection of the number of clusters, and adaptations for large scale applications. The performance of these improved approaches can be compared with the original approach.\\\\By addressing these issues, the authors can provide a more comprehensive evaluation of their method and its potential improvements.}

\agentmsg{Agent 0 (leader)}{magenta}{Step 3d: Write feedback comments for any identified shortcomings, explaining the issue and suggesting how the authors could resolve it.\\\\Based on the discussion with the agents, here are the feedback comments:\\\\1. **Ablation Studies**: The paper would benefit from ablation studies to evaluate the individual contributions of the novel components of the method, such as the probabilistic imbalance loss and the directed mixed path aggregation scheme (DIMPA). This could involve running experiments where one component is removed or modified at a time, and the impact on the results is observed. This would help to understand the importance of each component and how they contribute to the overall performance.\\\\2. **Incorporation of Label Information**: The paper mentions that the method can naturally incorporate node features and does not require label supervision. However, it would be beneficial to test the method's performance when additional label information is available. This could involve conducting experiments where a small amount of labeled data is used in addition to the unlabeled data, and the performance of the semi-supervised learning approach is compared with the unsupervised learning approach.\\\\3. **Potential Improvements**: The paper discusses potential improvements to the model, such as semi-supervised learning, automatic detection of the number of clusters, and adaptations for large scale applications. However, it would be beneficial to conduct experiments to evaluate these improvements. This would provide more concrete evidence of the method's potential and areas for future work.\\\\These comments should be addressed to provide a more comprehensive evaluation of the method and its potential improvements.}
\end{singlespace}

\subsection{MARG-S (clarity)}
\label{appendix:reviewgen_message_traces_margs_clarity}

\begin{singlespace}
\agentmsg{Agent 0 (leader)}{magenta}{High-level plan:\\\\1. Ask each agent (Agent 1, Agent 2, Agent 3) to summarize their part of the paper, focusing on the methods, experimental settings, and key concepts.\\2. Compile the summaries into a coherent explanation of the paper.\\3. Explain the paper to Agent 4 and answer any questions they have.\\4. Pass any questions or comments from Agent 4 to the relevant agent(s) (Agent 1, Agent 2, Agent 3) for clarification.\\5. Pass the answers back to Agent 4.\\6. Write feedback based on any points Agent 4 indicates are in need of improvement.\\7. Review and finalize the feedback comments.\\\\Now, let's execute the plan.\\\\Step 1: Ask each agent to summarize their part of the paper, focusing on the methods, experimental settings, and key concepts.\\\\SEND MESSAGE: Please provide a summary of your part of the paper, focusing on the methods, experimental settings, and key concepts.\\\\Expected response: Summaries from Agent 1, Agent 2, and Agent 3, each focusing on the methods, experimental settings, and key concepts in their part of the paper.}

\agentmsg{Agent 1 (worker)}{blue}{My part of the paper introduces a new Graph Neural Network (GNN) framework called DIGRAC, which is designed for clustering directed networks. Unlike existing methods that focus on edge density, DIGRAC uses directionality as the main signal for clustering. It introduces a novel probabilistic imbalance loss and optimizes directed flow imbalance for clustering without requiring label supervision. The paper also introduces a directed mixed path aggregation scheme, DIMPA, for the implementation of the framework. The method is tested on synthetic and real-world data, demonstrating superior performance in terms of the Adjusted Rand Index (ARI) and imbalance scores. The paper also discusses the applicability of DIGRAC in various fields such as time series analysis, ranking, and anomaly detection.\\\\Expected response: Confirmation from Agent 0 that the summary is clear and accurate, or requests for additional information or clarification.}

\agentmsg{Agent 2 (worker)}{agentyellow}{My part of the paper discusses a method for clustering in directed graphs, specifically focusing on the self-supervised loss for clustering and the directed mixed path aggregation (DIMPA) scheme. The method aims to find a partition that maximizes the imbalance flow to represent clusters in a meta-graph. The paper also introduces a global probabilistic imbalance score and discusses how to select pairs of clusters for this score. The DIMPA scheme is used to obtain the probability assignment matrix and feed it to the loss function. This scheme involves building node embeddings by capturing local network information and taking a weighted average of information from neighbors within a certain number of hops. The paper also discusses experiments conducted to test the method, using both synthetic and real-world data sets. The performance of the method is measured using the Adjusted Rand Index (ARI) and accuracy for node classification and link direction prediction. The paper also introduces normalized imbalance scores to evaluate clustering performance based on flow imbalance.\\\\Expected response: Confirmation from Agent 0 that the summary is accurate and complete, or requests for additional information or clarification.}

\agentmsg{Agent 3 (worker)}{agentgreen}{My part of the paper discusses the experimental results and future work of a model called DIGRAC. In the experimental results, the paper discusses how DIGRAC uses the largest K eigenvectors of a Hermitian matrix to construct an input feature matrix when node attributes are not available. The paper also discusses the use of different loss functions and the impact of adding seed nodes with a supervised loss function. In the future work section, the paper discusses the potential for DIGRAC to be used in a semi-supervised setting, the possibility of extending the framework to detect the number of clusters, and the need for preliminary analysis to determine how many pairwise imbalance scores to consider. The paper also mentions the potential for further research in the sparse regime and the adaptation of the pipeline for directed clustering in extremely large networks.\\\\EXPECTED RESPONSE: Acknowledgement from Agent 0 and possibly further questions or tasks related to the content of my part of the paper.}

\agentmsg{Agent 0 (leader)}{magenta}{Step 2: Compile the summaries into a coherent explanation of the paper.\\\\The paper introduces a new Graph Neural Network (GNN) framework called DIGRAC, designed for clustering directed networks. Unlike existing methods that focus on edge density, DIGRAC uses directionality as the main signal for clustering. It introduces a novel probabilistic imbalance loss and optimizes directed flow imbalance for clustering without requiring label supervision. The paper also introduces a directed mixed path aggregation scheme, DIMPA, for the implementation of the framework. The method is tested on synthetic and real-world data, demonstrating superior performance in terms of the Adjusted Rand Index (ARI) and imbalance scores. The paper also discusses the applicability of DIGRAC in various fields such as time series analysis, ranking, and anomaly detection.\\\\The method aims to find a partition that maximizes the imbalance flow to represent clusters in a meta-graph. The paper also introduces a global probabilistic imbalance score and discusses how to select pairs of clusters for this score. The DIMPA scheme is used to obtain the probability assignment matrix and feed it to the loss function. This scheme involves building node embeddings by capturing local network information and taking a weighted average of information from neighbors within a certain number of hops. The performance of the method is measured using the Adjusted Rand Index (ARI) and accuracy for node classification and link direction prediction. The paper also introduces normalized imbalance scores to evaluate clustering performance based on flow imbalance.\\\\In the experimental results, the paper discusses how DIGRAC uses the largest K eigenvectors of a Hermitian matrix to construct an input feature matrix when node attributes are not available. The paper also discusses the use of different loss functions and the impact of adding seed nodes with a supervised loss function. In the future work section, the paper discusses the potential for DIGRAC to be used in a semi-supervised setting, the possibility of extending the framework to detect the number of clusters, and the need for preliminary analysis to determine how many pairwise imbalance scores to consider. The paper also mentions the potential for further research in the sparse regime and the adaptation of the pipeline for directed clustering in extremely large networks.\\\\Step 3: Explain the paper to Agent 4 and answer any questions they have.\\\\SEND MESSAGE: I have summarized the paper based on the information provided by the other agents. The paper introduces a new Graph Neural Network (GNN) framework called DIGRAC, designed for clustering directed networks. It introduces a novel probabilistic imbalance loss and optimizes directed flow imbalance for clustering without requiring label supervision. The paper also introduces a directed mixed path aggregation scheme, DIMPA, for the implementation of the framework. The method is tested on synthetic and real-world data, demonstrating superior performance in terms of the Adjusted Rand Index (ARI) and imbalance scores. The paper also discusses the applicability of DIGRAC in various fields such as time series analysis, ranking, and anomaly detection. The method aims to find a partition that maximizes the imbalance flow to represent clusters in a meta-graph. The paper also introduces a global probabilistic imbalance score and discusses how to select pairs of clusters for this score. The DIMPA scheme is used to obtain the probability assignment matrix and feed it to the loss function. This scheme involves building node embeddings by capturing local network information and taking a weighted average of information from neighbors within a certain number of hops. The performance of the method is measured using the Adjusted Rand Index (ARI) and accuracy for node classification and link direction prediction. The paper also introduces normalized imbalance scores to evaluate clustering performance based on flow imbalance. In the experimental results, the paper discusses how DIGRAC uses the largest K eigenvectors of a Hermitian matrix to construct an input feature matrix when node attributes are not available. The paper also discusses the use of different loss functions and the impact of adding seed nodes with a supervised loss function. In the future work section, the paper discusses the potential for DIGRAC to be used in a semi-supervised setting, the possibility of extending the framework to detect the number of clusters, and the need for preliminary analysis to determine how many pairwise imbalance scores to consider. The paper also mentions the potential for further research in the sparse regime and the adaptation of the pipeline for directed clustering in extremely large networks. Do you have any questions or comments about the paper?\\\\Expected response: Questions or comments from Agent 4 about the paper.}

\agentmsg{Agent 4 (expert)}{cyan}{Thank you for the summary, Agent 0. I have several questions and comments about the paper:\\\\1. Could you provide more details about the novel probabilistic imbalance loss introduced in the paper? How is it calculated and what makes it different from other loss functions used in similar studies?\\\\2. The paper introduces a directed mixed path aggregation scheme, DIMPA. Could you elaborate on how this scheme works? What is the significance of taking a weighted average of information from neighbors within a certain number of hops?\\\\3. How does the paper define and calculate the Adjusted Rand Index (ARI) and accuracy for node classification and link direction prediction? \\\\4. The paper discusses the use of the largest K eigenvectors of a Hermitian matrix to construct an input feature matrix when node attributes are not available. Could you provide more details on this process? How is the value of K determined?\\\\5. The paper mentions the impact of adding seed nodes with a supervised loss function. Could you elaborate on this? How are these seed nodes selected and what is their role in the process?\\\\6. In the future work section, the paper discusses the potential for DIGRAC to be used in a semi-supervised setting. Could you provide more details on how this might be implemented?\\\\7. The paper also mentions the potential for further research in the sparse regime. Could you explain what is meant by the "sparse regime"?\\\\8. Lastly, could you provide more information on the experimental settings? What kind of synthetic and real-world data was used? What were the hyperparameters, equipment, and material specifications used in the experiments? \\\\These details are crucial for understanding the paper's methods, experimental settings, and key concepts, and for determining whether the paper is well-organized and can be easily understood and reproduced.}

\agentmsg{Agent 0 (leader)}{magenta}{Step 4: Pass any questions or comments from Agent 4 to the relevant agent(s) for clarification.\\\\SEND MESSAGE: Agent 4 has several questions about the paper. I would appreciate if you could provide answers to the following:\\\\1. Could you provide more details about the novel probabilistic imbalance loss introduced in the paper? How is it calculated and what makes it different from other loss functions used in similar studies?\\\\2. The paper introduces a directed mixed path aggregation scheme, DIMPA. Could you elaborate on how this scheme works? What is the significance of taking a weighted average of information from neighbors within a certain number of hops?\\\\3. How does the paper define and calculate the Adjusted Rand Index (ARI) and accuracy for node classification and link direction prediction? \\\\4. The paper discusses the use of the largest K eigenvectors of a Hermitian matrix to construct an input feature matrix when node attributes are not available. Could you provide more details on this process? How is the value of K determined?\\\\5. The paper mentions the impact of adding seed nodes with a supervised loss function. Could you elaborate on this? How are these seed nodes selected and what is their role in the process?\\\\6. In the future work section, the paper discusses the potential for DIGRAC to be used in a semi-supervised setting. Could you provide more details on how this might be implemented?\\\\7. The paper also mentions the potential for further research in the sparse regime. Could you explain what is meant by the "sparse regime"?\\\\8. Lastly, could you provide more information on the experimental settings? What kind of synthetic and real-world data was used? What were the hyperparameters, equipment, and material specifications used in the experiments?\\\\Expected response: Detailed answers to Agent 4's questions from Agent 1, Agent 2, and Agent 3.}

\agentmsg{Agent 1 (worker)}{blue}{In response to Agent 4's questions:\\\\1. The novel probabilistic imbalance loss introduced in the paper is a self-supervised loss function inspired by Cucuringu et al. (2020a). It aims to cluster the nodes by maximizing a normalized form of cut imbalance across clusters. The imbalance flows are normalized to obtain an imbalance score with values in [0, 1] for interpretability and ease of comparison.\\\\2. The directed mixed path aggregation scheme, DIMPA, is a method devised for the implementation of the DIGRAC framework. Unfortunately, my part of the paper does not provide detailed information on how this scheme works or the significance of taking a weighted average of information from neighbors within a certain number of hops.\\\\3. The Adjusted Rand Index (ARI) and accuracy for node classification and link direction prediction are mentioned as measures of performance in the paper. However, my part of the paper does not provide detailed information on how these are defined and calculated.\\\\4. The paper discusses the use of the largest K eigenvectors of a Hermitian matrix to construct an input feature matrix when node attributes are not available. Unfortunately, my part of the paper does not provide detailed information on this process or how the value of K is determined.\\\\5. The paper mentions the impact of adding seed nodes with a supervised loss function. However, my part of the paper does not provide detailed information on how these seed nodes are selected or their role in the process.\\\\6. In the future work section, the paper discusses the potential for DIGRAC to be used in a semi-supervised setting. However, my part of the paper does not provide detailed information on how this might be implemented.\\\\7. The term "sparse regime" is not explicitly defined in my part of the paper. However, in the context of graph theory, a sparse graph is one in which the number of edges is close to the minimal number of edges needed for the graph to be connected.\\\\8. My part of the paper does not provide detailed information on the experimental settings, the kind of synthetic and real-world data used, or the hyperparameters, equipment, and material specifications used in the experiments.\\\\I hope this helps. If you need more detailed information on any of these points, I would recommend reaching out to the other agents who may have access to different parts of the paper.\\\\Expected response: Confirmation from Agent 0 that the message has been received and understood, or follow-up questions for further clarification.}

\agentmsg{Agent 2 (worker)}{agentyellow}{In response to Agent 4's questions:\\\\1. The novel probabilistic imbalance loss introduced in the paper is designed to optimize the directed flow imbalance for clustering. It is calculated based on the CI vol\_sum from a specific equation mentioned in the paper. This loss function is different from others as it specifically targets the imbalance in the flow of information in the network, which is a unique aspect of directed networks.\\\\2. The Directed Mixed Path Aggregation (DIMPA) scheme is a method to obtain the probability assignment matrix and feed it to the loss function. It involves building node embeddings by capturing local network information and taking a weighted average of information from neighbors within a certain number of hops. This is significant as it allows the model to consider not just the immediate neighbors of a node, but also nodes that are further away, thus capturing more global information about the network.\\\\3. The Adjusted Rand Index (ARI) and accuracy for node classification and link direction prediction are standard metrics used in clustering and classification tasks. The paper does not provide specific details on how these are calculated.\\\\4. When node attributes are not available, the paper suggests using the largest K eigenvectors of a Hermitian matrix to construct an input feature matrix. The value of K is not specified in my section of the paper.\\\\5. The paper does mention the impact of adding seed nodes with a supervised loss function, but does not provide specific details on how these seed nodes are selected or their role in the process.\\\\6. The paper suggests that DIGRAC could potentially be used in a semi-supervised setting in future work, but does not provide specific details on how this might be implemented.\\\\7. The term "sparse regime" typically refers to situations where the data or network is sparse, i.e., there are fewer connections or interactions between nodes. The paper suggests that further research could be done in this area, but does not provide specific details.\\\\8. The paper uses both synthetic and real-world data for experiments. The synthetic data is generated using Directed Stochastic Block Models (DSBM), with various parameters such as the number of clusters, the ratio between the largest and smallest cluster size, and the number of nodes. The real-world data includes five datasets: Telegram, Blog, Migration, WikiTalk, and Lead-Lag. The paper does not provide specific details on the hyperparameters, equipment, or material specifications used in the experiments.\\\\I hope this helps answer Agent 4's questions. If there are any further queries, feel free to ask.\\\\Expected response: Confirmation from Agent 0 that the answers are satisfactory or further questions from Agent 4.}

\agentmsg{Agent 3 (worker)}{agentgreen}{I can provide answers to questions 4 and 5.\\\\4. The paper uses the largest K eigenvectors of a Hermitian matrix to construct an input feature matrix when node attributes are not available. This is a common technique in spectral graph theory, where the eigenvectors of a matrix associated with a graph (in this case, a Hermitian matrix) are used to extract features that capture the structure of the graph. The value of K is not specified in my section of the paper, and it may be determined based on the specific requirements of the application or through a process of experimentation.\\\\5. The paper discusses the impact of adding seed nodes with a supervised loss function. These seed nodes are nodes for which the cluster labels are known a priori, and they are used to guide the clustering process. A supervised loss function is applied to these seed nodes, which helps to improve the performance of the clustering algorithm. The specific method for selecting these seed nodes is not detailed in my section of the paper.\\\\I hope this helps. Please let me know if you need further clarification on these points.\\\\Expected response: Confirmation from Agent 0 that my answers have been received and are satisfactory, or follow-up questions for further clarification.}

\agentmsg{Agent 0 (leader)}{magenta}{Step 5: Pass the answers back to Agent 4.\\\\SEND MESSAGE: Here are the answers to your questions:\\\\1. The novel probabilistic imbalance loss introduced in the paper is a self-supervised loss function inspired by Cucuringu et al. (2020a). It aims to cluster the nodes by maximizing a normalized form of cut imbalance across clusters. The imbalance flows are normalized to obtain an imbalance score with values in [0, 1] for interpretability and ease of comparison. This loss function is designed to optimize the directed flow imbalance for clustering. It is calculated based on the CI vol\_sum from a specific equation mentioned in the paper. This loss function is different from others as it specifically targets the imbalance in the flow of information in the network, which is a unique aspect of directed networks.\\\\2. The Directed Mixed Path Aggregation (DIMPA) scheme is a method to obtain the probability assignment matrix and feed it to the loss function. It involves building node embeddings by capturing local network information and taking a weighted average of information from neighbors within a certain number of hops. This is significant as it allows the model to consider not just the immediate neighbors of a node, but also nodes that are further away, thus capturing more global information about the network.\\\\3. The Adjusted Rand Index (ARI) and accuracy for node classification and link direction prediction are standard metrics used in clustering and classification tasks. The paper does not provide specific details on how these are calculated.\\\\4. When node attributes are not available, the paper suggests using the largest K eigenvectors of a Hermitian matrix to construct an input feature matrix. This is a common technique in spectral graph theory, where the eigenvectors of a matrix associated with a graph (in this case, a Hermitian matrix) are used to extract features that capture the structure of the graph. The value of K is not specified in the paper, and it may be determined based on the specific requirements of the application or through a process of experimentation.\\\\5. The paper discusses the impact of adding seed nodes with a supervised loss function. These seed nodes are nodes for which the cluster labels are known a priori, and they are used to guide the clustering process. A supervised loss function is applied to these seed nodes, which helps to improve the performance of the clustering algorithm. The specific method for selecting these seed nodes is not detailed in the paper.\\\\6. The paper suggests that DIGRAC could potentially be used in a semi-supervised setting in future work, but does not provide specific details on how this might be implemented.\\\\7. The term "sparse regime" typically refers to situations where the data or network is sparse, i.e., there are fewer connections or interactions between nodes. The paper suggests that further research could be done in this area, but does not provide specific details.\\\\8. The paper uses both synthetic and real-world data for experiments. The synthetic data is generated using Directed Stochastic Block Models (DSBM), with various parameters such as the number of clusters, the ratio between the largest and smallest cluster size, and the number of nodes. The real-world data includes five datasets: Telegram, Blog, Migration, WikiTalk, and Lead-Lag. The paper does not provide specific details on the hyperparameters, equipment, or material specifications used in the experiments.\\\\Do you have any further questions or comments about the paper?\\\\Expected response: Further questions or comments from Agent 4 about the paper, or confirmation that they are finished.}

\agentmsg{Agent 4 (expert)}{cyan}{Thank you for the detailed responses, Agent 0. I have a few follow-up questions and comments:\\\\1. You mentioned that the novel probabilistic imbalance loss is calculated based on the CI vol\_sum from a specific equation mentioned in the paper. Could you provide more details about this equation? How is the CI vol\_sum calculated?\\\\2. Regarding the Directed Mixed Path Aggregation (DIMPA) scheme, you mentioned that it involves taking a weighted average of information from neighbors within a certain number of hops. How is this weighted average calculated? What factors determine the weight assigned to each neighbor?\\\\3. You mentioned that the paper does not provide specific details on how the Adjusted Rand Index (ARI) and accuracy for node classification and link direction prediction are calculated. This could be a potential issue as these metrics are crucial for evaluating the performance of the method. Could the authors provide more details on how these metrics are calculated?\\\\4. Regarding the use of the largest K eigenvectors of a Hermitian matrix to construct an input feature matrix, you mentioned that the value of K is not specified in the paper. This could be a potential issue as the choice of K could significantly impact the performance of the method. Could the authors provide more guidance on how to choose the value of K?\\\\5. You mentioned that the paper does not provide specific details on how seed nodes are selected for the supervised loss function. This could be a potential issue as the selection of seed nodes could significantly impact the performance of the method. Could the authors provide more guidance on how to select these seed nodes?\\\\6. You mentioned that the paper does not provide specific details on how DIGRAC could potentially be used in a semi-supervised setting. This could be a potential issue as it leaves the reader with unanswered questions about the future direction of the research. Could the authors provide more details on this?\\\\7. You mentioned that the paper does not provide specific details on the hyperparameters, equipment, or material specifications used in the experiments. This could be a potential issue as it makes it difficult for other researchers to reproduce the experiments. Could the authors provide more details on these aspects?\\\\These details are crucial for understanding the paper's methods, experimental settings, and key concepts, and for determining whether the paper is well-organized and can be easily understood and reproduced.}

\agentmsg{Agent 0 (leader)}{magenta}{Step 4 (Continued): Pass any follow-up questions or comments from Agent 4 to the relevant agent(s) for clarification.\\\\SEND MESSAGE: Agent 4 has several follow-up questions and comments about the paper. I would appreciate if you could provide further clarification on the following:\\\\1. Could you provide more details about the equation used to calculate the CI vol\_sum for the novel probabilistic imbalance loss? How is the CI vol\_sum calculated?\\\\2. Regarding the Directed Mixed Path Aggregation (DIMPA) scheme, how is the weighted average of information from neighbors within a certain number of hops calculated? What factors determine the weight assigned to each neighbor?\\\\3. Agent 4 noted that the paper does not provide specific details on how the Adjusted Rand Index (ARI) and accuracy for node classification and link direction prediction are calculated. Could you provide more details on how these metrics are calculated?\\\\4. Regarding the use of the largest K eigenvectors of a Hermitian matrix to construct an input feature matrix, Agent 4 noted that the value of K is not specified in the paper. Could you provide more guidance on how to choose the value of K?\\\\5. Agent 4 noted that the paper does not provide specific details on how seed nodes are selected for the supervised loss function. Could you provide more guidance on how to select these seed nodes?\\\\6. Agent 4 noted that the paper does not provide specific details on how DIGRAC could potentially be used in a semi-supervised setting. Could you provide more details on this?\\\\7. Agent 4 noted that the paper does not provide specific details on the hyperparameters, equipment, or material specifications used in the experiments. Could you provide more details on these aspects?\\\\Expected response: Detailed answers to Agent 4's follow-up questions from Agent 1, Agent 2, and Agent 3.}

\agentmsg{Agent 2 (worker)}{agentyellow}{Regarding the Directed Mixed Path Aggregation (DIMPA) scheme, the weighted average of information from neighbors within a certain number of hops is calculated by row-normalizing the adjacency matrix to obtain A s. This is then used to capture local network information. The weight assigned to each neighbor is determined by the adjacency matrix and the number of hops. The paper does not provide specific details on the calculation of these weights.\\\\As for the use of the largest K eigenvectors of a Hermitian matrix to construct an input feature matrix when node attributes are not available, the value of K is not specified in the paper. It may be determined based on the specific requirements of the application or through a process of experimentation.\\\\Expected response: Confirmation of the information provided or further questions on the DIMPA scheme and the use of eigenvectors.}

\agentmsg{Agent 3 (worker)}{agentgreen}{In response to Agent 4's questions:\\\\1. The CI vol\_sum is not explicitly defined in my section of the paper. It seems to be related to the loss function used in the DIGRAC model, but without further context or information, I cannot provide a detailed explanation of how it is calculated.\\\\2. The Directed Mixed Path Aggregation (DIMPA) scheme is not detailed in my section. I would need more information to provide a detailed explanation of how the weighted average of information from neighbors is calculated.\\\\3. The Adjusted Rand Index (ARI) and accuracy for node classification and link direction prediction are not explained in my section. These are standard metrics in machine learning and their calculation would be based on standard definitions.\\\\4. The paper suggests using the largest K eigenvectors of a Hermitian matrix to construct an input feature matrix when node attributes are not available. The value of K is not specified in my section of the paper. It may be determined based on the specific requirements of the application or through a process of experimentation.\\\\5. The paper discusses the impact of adding seed nodes with a supervised loss function. These seed nodes are nodes for which the cluster labels are known a priori, and they are used to guide the clustering process. The specific method for selecting these seed nodes is not detailed in my section of the paper.\\\\6. The paper suggests that DIGRAC could potentially be used in a semi-supervised setting in future work, but does not provide specific details on how this might be implemented in my section.\\\\7. The paper uses both synthetic and real-world data for experiments. The specific details on the hyperparameters, equipment, or material specifications used in the experiments are not provided in my section of the paper.\\\\I expect the response to contain further questions or clarifications from Agent 0 or Agent 4, or instructions for the next steps.}

\agentmsg{Agent 0 (leader)}{magenta}{Step 5 (Continued): Pass the answers back to Agent 4.\\\\SEND MESSAGE: Here are the further clarifications to your follow-up questions:\\\\1. The CI vol\_sum is related to the loss function used in the DIGRAC model, but the agents do not have further context or information to provide a detailed explanation of how it is calculated.\\\\2. The weighted average of information from neighbors within a certain number of hops in the Directed Mixed Path Aggregation (DIMPA) scheme is calculated by row-normalizing the adjacency matrix to obtain A s. This is then used to capture local network information. The weight assigned to each neighbor is determined by the adjacency matrix and the number of hops. The paper does not provide specific details on the calculation of these weights.\\\\3. The Adjusted Rand Index (ARI) and accuracy for node classification and link direction prediction are standard metrics in machine learning and their calculation would be based on standard definitions. The paper does not provide specific details on how these are calculated.\\\\4. When node attributes are not available, the paper suggests using the largest K eigenvectors of a Hermitian matrix to construct an input feature matrix. The value of K is not specified in the paper. It may be determined based on the specific requirements of the application or through a process of experimentation.\\\\5. The paper discusses the impact of adding seed nodes with a supervised loss function. These seed nodes are nodes for which the cluster labels are known a priori, and they are used to guide the clustering process. The specific method for selecting these seed nodes is not detailed in the paper.\\\\6. The paper suggests that DIGRAC could potentially be used in a semi-supervised setting in future work, but does not provide specific details on how this might be implemented.\\\\7. The paper uses both synthetic and real-world data for experiments. The specific details on the hyperparameters, equipment, or material specifications used in the experiments are not provided in the paper.\\\\Do you have any further questions or comments about the paper?\\\\Expected response: Further questions or comments from Agent 4 about the paper, or confirmation that they are finished.}

\agentmsg{Agent 4 (expert)}{cyan}{Thank you for your responses, Agent 0. I have no further questions at this time. However, I would like to summarize the potential issues and suggestions for the paper:\\\\1. The paper should provide more details about the novel probabilistic imbalance loss, specifically how the CI vol\_sum is calculated in the equation mentioned.\\\\2. The paper should provide more details about the Directed Mixed Path Aggregation (DIMPA) scheme, specifically how the weighted average of information from neighbors is calculated and what factors determine the weight assigned to each neighbor.\\\\3. The paper should provide more details on how the Adjusted Rand Index (ARI) and accuracy for node classification and link direction prediction are calculated.\\\\4. The paper should provide more guidance on how to choose the value of K when using the largest K eigenvectors of a Hermitian matrix to construct an input feature matrix.\\\\5. The paper should provide more guidance on how to select seed nodes for the supervised loss function.\\\\6. The paper should provide more details on how DIGRAC could potentially be used in a semi-supervised setting.\\\\7. The paper should provide more details on the hyperparameters, equipment, and material specifications used in the experiments to make it easier for other researchers to reproduce the experiments.\\\\These suggestions are aimed at improving the clarity and reproducibility of the paper. I am now done with my review.}

\agentmsg{Agent 0 (leader)}{magenta}{Step 6: Write feedback based on any points Agent 4 indicates are in need of improvement.\\\\Based on Agent 4's comments, here is the list of feedback comments for the paper:\\\\1. The paper should provide more details about the novel probabilistic imbalance loss, specifically how the CI vol\_sum is calculated in the equation mentioned.\\\\2. The paper should provide more details about the Directed Mixed Path Aggregation (DIMPA) scheme, specifically how the weighted average of information from neighbors is calculated and what factors determine the weight assigned to each neighbor.\\\\3. The paper should provide more details on how the Adjusted Rand Index (ARI) and accuracy for node classification and link direction prediction are calculated.\\\\4. The paper should provide more guidance on how to choose the value of K when using the largest K eigenvectors of a Hermitian matrix to construct an input feature matrix.\\\\5. The paper should provide more guidance on how to select seed nodes for the supervised loss function.\\\\6. The paper should provide more details on how DIGRAC could potentially be used in a semi-supervised setting.\\\\7. The paper should provide more details on the hyperparameters, equipment, and material specifications used in the experiments to make it easier for other researchers to reproduce the experiments.\\\\Step 7: Review and finalize the feedback comments.\\\\The feedback comments provided above are based on the comments from Agent 4 and are aimed at improving the clarity and reproducibility of the paper. These comments are now finalized.}
\end{singlespace}

\subsection{MARG-S (impact)}
\label{appendix:reviewgen_message_traces_margs_impact}

\begin{singlespace}
\agentmsg{Agent 0 (leader)}{magenta}{High-level plan:\\\\1. Ask each agent (1, 2, 3) to summarize their part of the paper, focusing on the motivations, goals, and key findings.\\2. Compile the summaries into a coherent explanation of the paper.\\3. Share the explanation with Agent 4 and ask for their questions and potential shortcomings.\\4. Relay Agent 4's questions to the relevant agents (1, 2, 3) and get their responses.\\5. Share the responses with Agent 4 and continue the discussion until Agent 4 is finished.\\6. Based on the discussion with Agent 4 and the information from the other agents, write a list of feedback comments.\\\\Now, let's execute the plan.\\\\Step 1: Ask each agent (1, 2, 3) to summarize their part of the paper, focusing on the motivations, goals, and key findings.\\\\SEND MESSAGE: Please provide a summary of your part of the paper, focusing on the motivations, goals, and key findings.\\\\Expected response: Summaries from each agent that include the motivations, goals, and key findings of their part of the paper.}

\agentmsg{Agent 1 (worker)}{blue}{The part of the paper I have focuses on introducing a new method for node clustering in directed networks, called DIGRAC. The motivation behind this method is the observation that directionality in networks can reveal latent properties of network flows and can be a vital signal for directed clustering, even in the absence of any edge density differences. The goal of DIGRAC is to recover clusters with strong and imbalanced flow among them. The key findings presented in this part of the paper are that DIGRAC, which optimizes directed flow imbalance for clustering without requiring label supervision, attains state-of-the-art results on directed graph clustering for a wide range of noise and sparsity levels and graph structures and topologies. The paper also discusses how DIGRAC complements existing approaches and extends its applicability beyond settings where the input data is a digraph.\\\\Expected response: Confirmation from Agent 0 that the summary is accurate and complete, or requests for additional information or clarification.}

\agentmsg{Agent 2 (worker)}{agentyellow}{My part of the paper primarily discusses the methodology and experimental results of a novel method for clustering in directed graphs, called DIGRAC. The method uses a self-supervised loss function for clustering and a directed mixed path aggregation scheme to build node embeddings. The paper also introduces a new measure, the imbalance score, to evaluate clustering performance based on flow imbalance. The experiments section compares DIGRAC against other methods from the literature and shows that DIGRAC performs well across a range of network densities and noise levels, and is particularly effective when there are ambient nodes present. \\\\Expected response: Confirmation that the summary is clear and accurate, or requests for additional information or clarification.}

\agentmsg{Agent 3 (worker)}{agentgreen}{My part of the paper primarily covers the experimental results and conclusions of a study on DIGRAC, a model for node clustering. The experimental results section discusses the performance of DIGRAC on real-world data sets, with a focus on the impact of different loss functions and the addition of seed nodes. The conclusion and future work sections outline the current capabilities of DIGRAC, such as creating node embeddings and performing directed clustering, and suggest future improvements, including semi-supervised settings, automatic detection of the number of clusters, and adaptations for large scale industrial applications.\\\\Expected response: Confirmation from Agent 0 that my summary is clear and accurate, or requests for additional details or clarifications.}

\agentmsg{Agent 0 (leader)}{magenta}{Step 2: Compile the summaries into a coherent explanation of the paper.\\\\The paper introduces a new method for node clustering in directed networks, called DIGRAC. The motivation behind this method is the observation that directionality in networks can reveal latent properties of network flows and can be a vital signal for directed clustering, even in the absence of any edge density differences. The goal of DIGRAC is to recover clusters with strong and imbalanced flow among them. The method uses a self-supervised loss function for clustering and a directed mixed path aggregation scheme to build node embeddings. The paper also introduces a new measure, the imbalance score, to evaluate clustering performance based on flow imbalance. The key findings presented in the paper are that DIGRAC, which optimizes directed flow imbalance for clustering without requiring label supervision, attains state-of-the-art results on directed graph clustering for a wide range of noise and sparsity levels and graph structures and topologies. The paper also discusses how DIGRAC complements existing approaches and extends its applicability beyond settings where the input data is a digraph. The experimental results section compares DIGRAC against other methods from the literature and shows that DIGRAC performs well across a range of network densities and noise levels, and is particularly effective when there are ambient nodes present. The conclusion and future work sections outline the current capabilities of DIGRAC, such as creating node embeddings and performing directed clustering, and suggest future improvements, including semi-supervised settings, automatic detection of the number of clusters, and adaptations for large scale industrial applications.\\\\Step 3: Share the explanation with Agent 4 and ask for their questions and potential shortcomings.\\\\SEND MESSAGE: Here is a summary of the paper: [insert summary here]. Do you have any questions or potential shortcomings to point out?\\\\Expected response: Questions and potential shortcomings from Agent 4.}

\agentmsg{Agent 4 (expert)}{cyan}{I'm sorry, but there seems to be a mistake. The summary of the paper was not included in your message. Could you please provide the summary so I can start my review?}

\agentmsg{Agent 0 (leader)}{magenta}{Apologies for the oversight. Let's correct that.\\\\SEND MESSAGE: Here is a summary of the paper: The paper introduces a new method for node clustering in directed networks, called DIGRAC. The motivation behind this method is the observation that directionality in networks can reveal latent properties of network flows and can be a vital signal for directed clustering, even in the absence of any edge density differences. The goal of DIGRAC is to recover clusters with strong and imbalanced flow among them. The method uses a self-supervised loss function for clustering and a directed mixed path aggregation scheme to build node embeddings. The paper also introduces a new measure, the imbalance score, to evaluate clustering performance based on flow imbalance. The key findings presented in the paper are that DIGRAC, which optimizes directed flow imbalance for clustering without requiring label supervision, attains state-of-the-art results on directed graph clustering for a wide range of noise and sparsity levels and graph structures and topologies. The paper also discusses how DIGRAC complements existing approaches and extends its applicability beyond settings where the input data is a digraph. The experimental results section compares DIGRAC against other methods from the literature and shows that DIGRAC performs well across a range of network densities and noise levels, and is particularly effective when there are ambient nodes present. The conclusion and future work sections outline the current capabilities of DIGRAC, such as creating node embeddings and performing directed clustering, and suggest future improvements, including semi-supervised settings, automatic detection of the number of clusters, and adaptations for large scale industrial applications. Do you have any questions or potential shortcomings to point out?\\\\Expected response: Questions and potential shortcomings from Agent 4.}

\agentmsg{Agent 3 (worker)}{agentgreen}{I have a question regarding the "imbalance score" mentioned in the summary. My part of the paper does not provide a definition or explanation for this term. Could you please provide more information about what the imbalance score is and how it is used in the DIGRAC method?\\\\I expect the response to provide a detailed explanation of the imbalance score, including its definition and its role in the DIGRAC method.}

\agentmsg{Agent 4 (expert)}{cyan}{Thank you for the summary. I have several questions and potential shortcomings to point out:\\\\1. The paper introduces a new measure, the imbalance score, to evaluate clustering performance based on flow imbalance. How is this imbalance score calculated? Is it a standard measure in the field or a novel contribution of this paper? If it's a novel contribution, how do the authors justify its use and effectiveness?\\\\2. The paper claims that DIGRAC attains state-of-the-art results on directed graph clustering for a wide range of noise and sparsity levels and graph structures and topologies. What specific benchmarks or datasets were used to validate this claim? How does DIGRAC compare to existing methods in terms of computational efficiency?\\\\3. The paper mentions that DIGRAC is particularly effective when there are ambient nodes present. Could you clarify what is meant by "ambient nodes"? \\\\4. The future work section suggests adaptations for large scale industrial applications. Does the paper provide any preliminary results or discussions on how DIGRAC might perform in such settings? \\\\5. The paper discusses how DIGRAC complements existing approaches and extends its applicability beyond settings where the input data is a digraph. Could you provide more details on how this is achieved? \\\\6. The paper introduces a self-supervised loss function for clustering and a directed mixed path aggregation scheme to build node embeddings. Are there any assumptions or limitations associated with these methods that the authors should address?\\\\7. Lastly, the paper suggests future improvements, including semi-supervised settings, automatic detection of the number of clusters. Are there any existing methods that already incorporate these features? If so, how does DIGRAC plan to improve upon them?}

\agentmsg{Agent 0 (leader)}{magenta}{Step 4: Relay Agent 4's questions to the relevant agents (1, 2, 3) and get their responses.\\\\SEND MESSAGE: Agent 4 has several questions about the paper. I'll list them below. Please provide answers to any questions that pertain to your section of the paper.\\\\1. How is the imbalance score calculated? Is it a standard measure in the field or a novel contribution of this paper? If it's a novel contribution, how do the authors justify its use and effectiveness?\\2. What specific benchmarks or datasets were used to validate the claim that DIGRAC attains state-of-the-art results on directed graph clustering for a wide range of noise and sparsity levels and graph structures and topologies? How does DIGRAC compare to existing methods in terms of computational efficiency?\\3. Could you clarify what is meant by "ambient nodes"?\\4. Does the paper provide any preliminary results or discussions on how DIGRAC might perform in large scale industrial applications?\\5. Could you provide more details on how DIGRAC complements existing approaches and extends its applicability beyond settings where the input data is a digraph?\\6. Are there any assumptions or limitations associated with the self-supervised loss function for clustering and the directed mixed path aggregation scheme to build node embeddings?\\7. Are there any existing methods that already incorporate semi-supervised settings and automatic detection of the number of clusters? If so, how does DIGRAC plan to improve upon them?\\\\Expected response: Answers to Agent 4's questions from the relevant agents.}

\agentmsg{Agent 1 (worker)}{blue}{In response to Agent 4's questions:\\\\1. The imbalance score is a novel contribution of this paper. It is calculated as a normalized form of cut imbalance across clusters. The authors justify its use and effectiveness through experimental results on synthetic data and real-world data at different scales.\\\\2. The paper does not provide specific benchmarks or datasets in my section. However, it mentions that DIGRAC attains state-of-the-art results on directed graph clustering for a wide range of noise and sparsity levels and graph structures and topologies.\\\\3. The term "ambient nodes" is not mentioned in my section of the paper.\\\\4. The paper does not provide any preliminary results or discussions on how DIGRAC might perform in large scale industrial applications in my section.\\\\5. DIGRAC complements existing approaches in various aspects. It can detect alternative patterns in the data, such as meta-graph structures, which are otherwise not detectable by existing methods. It can include exogenous information, in the form of node-level features or labels, thus borrowing their strength. It also introduces an imbalance-based objective.\\\\6. The self-supervised loss function for clustering is inspired by Cucuringu et al. (2020a), aiming to cluster the nodes by maximizing a normalized form of cut imbalance across clusters. The directed mixed path aggregation scheme, abbreviated as DIMPA, is devised for the implementation of the framework. It is described as simple and effective.\\\\7. The paper does not provide information on existing methods that already incorporate semi-supervised settings and automatic detection of the number of clusters in my section.\\\\I expect Agent 0 to confirm the accuracy of my responses or provide additional information if necessary.}

\agentmsg{Agent 2 (worker)}{agentyellow}{In response to question 1, the imbalance score is calculated based on the concept of "imbalance flow" which is introduced in this paper. It is a measure of how much the flow of information in a network deviates from a balanced state. The imbalance score is a novel contribution of this paper and its effectiveness is justified through the experimental results presented in the paper.\\\\In response to question 2, the paper uses both synthetic and real-world datasets to validate the performance of DIGRAC. The synthetic datasets are generated using Directed Stochastic Block Models with varying parameters to simulate different network densities and noise levels. The real-world datasets include Telegram, Blog, Migration, WikiTalk, and Lead-Lag. The paper does not provide a direct comparison of computational efficiency between DIGRAC and existing methods.\\\\In response to question 3, "ambient nodes" refer to nodes that are not part of a meaningful cluster in the network. They are not part of the main directional signal planted in the graph and are considered as background noise.\\\\In response to question 6, the self-supervised loss function for clustering and the directed mixed path aggregation scheme to build node embeddings are novel contributions of this paper. The loss function is designed to maximize the imbalance flow in the network, which is a unique approach compared to traditional clustering methods that aim to maximize intra-cluster similarity and inter-cluster dissimilarity. The directed mixed path aggregation scheme is used to capture local network information by taking a weighted average of information from neighbors within a certain number of hops. The paper does not discuss any specific assumptions or limitations associated with these methods.\\\\EXPECTED RESPONSE: Confirmation from Agent 0 that the answers provided are satisfactory or further questions from Agent 4.}

\agentmsg{Agent 3 (worker)}{agentgreen}{I can provide some information regarding questions 2 and 4. \\\\For question 2, my section of the paper discusses the experimental results of DIGRAC on real-world data sets. The paper mentions that the node in-and outdegrees may not be identical across clusters in these data sets. The authors used the largest K eigenvectors of the Hermitian matrix to construct an input feature matrix. They experimented with different loss functions and found that the "sort" variant performed best. However, the specific benchmarks or datasets used for these experiments are not mentioned in my section.\\\\For question 4, the conclusion of my section mentions that adapting DIGRAC for directed clustering in extremely large networks, possibly combined with sampling methods or mini-batch, is a future direction worth exploring. This suggests that the authors are considering the application of DIGRAC in large scale industrial applications, but no preliminary results or discussions are provided in my section.\\\\I expect the response to acknowledge the information I provided and possibly provide additional details if available from other sections of the paper.}

\agentmsg{Agent 0 (leader)}{magenta}{Step 5: Share the responses with Agent 4 and continue the discussion until Agent 4 is finished.\\\\SEND MESSAGE: Here are the responses to your questions:\\\\1. The imbalance score is a novel contribution of this paper. It is calculated as a normalized form of cut imbalance across clusters. The authors justify its use and effectiveness through experimental results on synthetic data and real-world data at different scales.\\2. The paper uses both synthetic and real-world datasets to validate the performance of DIGRAC. The synthetic datasets are generated using Directed Stochastic Block Models with varying parameters to simulate different network densities and noise levels. The real-world datasets include Telegram, Blog, Migration, WikiTalk, and Lead-Lag. The paper does not provide a direct comparison of computational efficiency between DIGRAC and existing methods.\\3. "Ambient nodes" refer to nodes that are not part of a meaningful cluster in the network. They are not part of the main directional signal planted in the graph and are considered as background noise.\\4. The conclusion of the paper mentions that adapting DIGRAC for directed clustering in extremely large networks, possibly combined with sampling methods or mini-batch, is a future direction worth exploring. This suggests that the authors are considering the application of DIGRAC in large scale industrial applications, but no preliminary results or discussions are provided.\\5. DIGRAC complements existing approaches in various aspects. It can detect alternative patterns in the data, such as meta-graph structures, which are otherwise not detectable by existing methods. It can include exogenous information, in the form of node-level features or labels, thus borrowing their strength. It also introduces an imbalance-based objective.\\6. The self-supervised loss function for clustering and the directed mixed path aggregation scheme to build node embeddings are novel contributions of this paper. The loss function is designed to maximize the imbalance flow in the network, which is a unique approach compared to traditional clustering methods that aim to maximize intra-cluster similarity and inter-cluster dissimilarity. The directed mixed path aggregation scheme is used to capture local network information by taking a weighted average of information from neighbors within a certain number of hops. The paper does not discuss any specific assumptions or limitations associated with these methods.\\7. The paper does not provide information on existing methods that already incorporate semi-supervised settings and automatic detection of the number of clusters.\\\\Do you have any further questions or comments?\\\\Expected response: Further questions or comments from Agent 4, or a confirmation that they are finished.}

\agentmsg{Agent 4 (expert)}{cyan}{Thank you for the detailed responses. I have a few follow-up questions and comments:\\\\1. Regarding the imbalance score, it's good to know that it's a novel contribution of this paper. However, since it's a new measure, it would be beneficial if the authors could compare it with other existing measures in the field. This would help to validate its effectiveness and provide a more comprehensive evaluation of DIGRAC's performance.\\\\2. It's concerning that the paper does not provide a direct comparison of computational efficiency between DIGRAC and existing methods. This is an important aspect to consider, especially when discussing the potential for large scale industrial applications. The authors should provide this information to give a complete picture of DIGRAC's performance.\\\\3. The explanation of "ambient nodes" is clear. However, it would be interesting to know how DIGRAC handles these nodes. Are they simply ignored or does the algorithm have a specific way of dealing with them?\\\\4. The authors should provide more details on how DIGRAC can detect alternative patterns in the data, such as meta-graph structures, and how it can include exogenous information. These are important features that could set DIGRAC apart from existing methods, so they should be explained more thoroughly.\\\\5. The self-supervised loss function and the directed mixed path aggregation scheme sound interesting. However, the authors should discuss any potential limitations or assumptions associated with these methods. This would help to identify any potential shortcomings and areas for future improvement.\\\\6. The lack of information on existing methods that incorporate semi-supervised settings and automatic detection of the number of clusters is a missed opportunity. The authors should provide a comparison with these methods to show how DIGRAC could potentially improve upon them.\\\\In summary, while the paper introduces some novel concepts and methods, there are several areas where more information or justification is needed. The authors should provide more details on the imbalance score, computational efficiency, handling of ambient nodes, detection of alternative patterns, inclusion of exogenous information, and potential limitations of the proposed methods. They should also compare DIGRAC with existing methods that incorporate semi-supervised settings and automatic detection of the number of clusters.}

\agentmsg{Agent 0 (leader)}{magenta}{Step 6: Based on the discussion with Agent 4 and the information from the other agents, write a list of feedback comments.\\\\Feedback comments:\\\\1. The imbalance score is a novel contribution of this paper. However, since it's a new measure, it would be beneficial if the authors could compare it with other existing measures in the field. This would help to validate its effectiveness and provide a more comprehensive evaluation of DIGRAC's performance.\\\\2. The paper does not provide a direct comparison of computational efficiency between DIGRAC and existing methods. This is an important aspect to consider, especially when discussing the potential for large scale industrial applications. The authors should provide this information to give a complete picture of DIGRAC's performance.\\\\3. It would be interesting to know how DIGRAC handles "ambient nodes". Are they simply ignored or does the algorithm have a specific way of dealing with them?\\\\4. The authors should provide more details on how DIGRAC can detect alternative patterns in the data, such as meta-graph structures, and how it can include exogenous information. These are important features that could set DIGRAC apart from existing methods, so they should be explained more thoroughly.\\\\5. The authors should discuss any potential limitations or assumptions associated with the self-supervised loss function and the directed mixed path aggregation scheme. This would help to identify any potential shortcomings and areas for future improvement.\\\\6. The lack of information on existing methods that incorporate semi-supervised settings and automatic detection of the number of clusters is a missed opportunity. The authors should provide a comparison with these methods to show how DIGRAC could potentially improve upon them.\\\\In summary, while the paper introduces some novel concepts and methods, there are several areas where more information or justification is needed. The authors should provide more details on the imbalance score, computational efficiency, handling of ambient nodes, detection of alternative patterns, inclusion of exogenous information, and potential limitations of the proposed methods. They should also compare DIGRAC with existing methods that incorporate semi-supervised settings and automatic detection of the number of clusters.}
\end{singlespace}

\subsection{Refinement Stage}
\label{appendix:reviewgen_message_traces_refinement1}

The comment being refined is as follows:

\begin{displayquote}
    \begin{singlespace}
        \reviewtext{Ablation Studies: The paper would benefit from ablation studies to evaluate the individual contributions of the novel components of the method, such as the probabilistic imbalance loss and the directed mixed path aggregation scheme (DIMPA). This could involve running experiments where one component is removed or modified at a time, and the impact on the results is observed. This would help to understand the importance of each component and how they contribute to the overall performance.}
    \end{singlespace}
\end{displayquote}

\begin{singlespace}
\agentmsg{Agent 0 (leader)}{magenta}{High-level plan:\\\\1. Understand the comment: The comment is about the lack of ablation studies in the paper. Ablation studies are used to understand the contribution of each component of a method by removing or modifying them one at a time. The comment suggests that the paper would benefit from such studies, particularly for the novel components like the probabilistic imbalance loss and the directed mixed path aggregation scheme (DIMPA).\\\\2. Share the comment with other agents: I will share the comment with the other agents and ask them to check their parts of the paper to see if there are any ablation studies mentioned. \\\\3. Gather information: I will ask the agents to provide information about the novel components mentioned in the comment (probabilistic imbalance loss and DIMPA), their role in the method, and any results or discussions related to them in the paper.\\\\4. Evaluate the comment: Based on the information gathered, I will evaluate whether the comment is valid and important. If the paper already includes ablation studies or if the novel components are sufficiently explained and justified, the comment may not be necessary. \\\\5. Refine the comment: If the comment is valid, I will refine it to be more specific and helpful. This could involve suggesting specific experiments for the ablation studies or providing more context about why these studies are important.\\\\6. Finalize the comment: I will finalize the comment and ensure it meets the criteria of being detailed, helpful, and important.\\\\Now, let's execute the plan.\\\\Step 1: Understanding the comment\\The comment is suggesting that the paper lacks ablation studies, particularly for the novel components like the probabilistic imbalance loss and the directed mixed path aggregation scheme (DIMPA). These studies could help understand the importance of each component and how they contribute to the overall performance.\\\\Step 2: Sharing the comment with other agents\\SEND MESSAGE: We have received a review comment suggesting that the paper lacks ablation studies for the novel components of the method, such as the probabilistic imbalance loss and the directed mixed path aggregation scheme (DIMPA). Please check your parts of the paper to see if there are any ablation studies mentioned. Also, provide information about these novel components, their role in the method, and any results or discussions related to them.\\\\Expected response: The agents should provide information about whether there are any ablation studies in their parts of the paper and details about the novel components mentioned in the comment.}

\agentmsg{Agent 1 (worker)}{blue}{In my part of the paper, the authors introduce a novel probabilistic imbalance loss and a directed mixed path aggregation scheme (DIMPA) as part of their proposed method, DIGRAC. The probabilistic imbalance loss is used in a self-supervised manner to act on the digraph induced by all training nodes. The global imbalance score, one minus whom is the self-supervised loss function, is aggregated from pairwise normalized cut imbalances. The method is end-to-end in combining embedding generation and clustering without an intermediate step. The directed mixed path aggregation scheme (DIMPA) is part of the implementation of the framework. However, there is no explicit mention of ablation studies for these components in my part of the paper.\\\\Expected response: Confirmation from Agent 0 about the information provided and possibly further instructions.}

\agentmsg{Agent 2 (worker)}{agentyellow}{In my part of the paper, the authors discuss the novel components of the method, namely the probabilistic imbalance loss and the directed mixed path aggregation scheme (DIMPA). \\\\The probabilistic imbalance loss is used to find a partition that maximizes the imbalance flow, capturing groups of nodes that represent clusters in the meta-graph. The normalization by volumes is carried out to penalize partitions that put most nodes into a single cluster. The authors also discuss different variants of this loss function.\\\\The DIMPA scheme is used to implement DIGRAC. It is a digraph extension of the KernelGCN method and is used to build node embeddings by capturing local network information. The authors describe how they row-normalize the adjacency matrix and add a weighted self-loop to each node. They also discuss how they define two feature mapping functions for source and target embeddings.\\\\However, there is no explicit mention of ablation studies for these components in my part of the paper. The authors do discuss the efficacy of their method in the experiments section, but they do not isolate the impact of these individual components. \\\\EXPECTED RESPONSE: Agent 0 should confirm receipt of this information and may provide further instructions or questions. They may also ask other agents for additional information if needed.}

\agentmsg{Agent 3 (worker)}{agentgreen}{In my part of the paper, there is no explicit mention of ablation studies for the probabilistic imbalance loss or the directed mixed path aggregation scheme (DIMPA). However, there is a discussion about the performance of different loss functions, including a supervised loss function and a variant called L sort vol\_sum. The paper suggests that adding L sort vol\_sum to the loss could be beneficial and that the current choice "sort" performs best among these variants. There is no mention of a probabilistic imbalance loss or DIMPA in my section.\\\\Expected response: Agent 0 should confirm receipt of my message and may provide further instructions or questions based on the information I provided. They may also share information from other agents regarding the ablation studies and the novel components of the method.}

\agentmsg{Agent 0 (leader)}{magenta}{Step 3: Gathering information\\The agents have confirmed that there are no explicit ablation studies mentioned in their parts of the paper. They have also provided detailed information about the novel components of the method, the probabilistic imbalance loss and the directed mixed path aggregation scheme (DIMPA). \\\\Step 4: Evaluating the comment\\Based on the information provided by the agents, it seems that the comment is valid. The paper introduces novel components but does not conduct ablation studies to evaluate their individual contributions. This could make it difficult for readers to understand the importance of each component and how they contribute to the overall performance.\\\\Step 5: Refining the comment\\Now, I will refine the comment to make it more specific and helpful. I will suggest specific experiments for the ablation studies and provide more context about why these studies are important.\\\\SEND MESSAGE: The paper introduces a novel probabilistic imbalance loss and a directed mixed path aggregation scheme (DIMPA) as part of the proposed method. However, there are no ablation studies to evaluate their individual contributions. It would be beneficial to conduct experiments where the probabilistic imbalance loss and DIMPA are removed or modified one at a time, and the impact on the results is observed. This would help to understand the importance of each component and how they contribute to the overall performance. For example, an ablation study could involve comparing the performance of the method with and without the probabilistic imbalance loss, or with different variants of this loss function. Similarly, the impact of the DIMPA scheme could be evaluated by comparing the performance with other path aggregation schemes.\\\\Expected response: The agents should confirm receipt of the refined comment and provide any additional feedback or suggestions. \\\\Step 6: Finalizing the comment\\Once the agents confirm the refined comment, I will finalize it. The final comment should be detailed, helpful, and important, providing clear guidance for the authors on how to improve their paper.}

\agentmsg{Agent 0 (leader)}{magenta}{Apologies for the oversight. Let's correct that.\\\\SEND MESSAGE TO ALL AGENTS: The paper introduces a novel probabilistic imbalance loss and a directed mixed path aggregation scheme (DIMPA) as part of the proposed method. However, there are no ablation studies to evaluate their individual contributions. It would be beneficial to conduct experiments where the probabilistic imbalance loss and DIMPA are removed or modified one at a time, and the impact on the results is observed. This would help to understand the importance of each component and how they contribute to the overall performance. For example, an ablation study could involve comparing the performance of the method with and without the probabilistic imbalance loss, or with different variants of this loss function. Similarly, the impact of the DIMPA scheme could be evaluated by comparing the performance with other path aggregation schemes. Please confirm receipt of this refined comment and provide any additional feedback or suggestions.\\\\Expected response: The agents should confirm receipt of the refined comment and provide any additional feedback or suggestions.}

\agentmsg{Agent 1 (worker)}{blue}{I confirm receipt of the refined comment. I agree that an ablation study would be beneficial to understand the individual contributions of the novel probabilistic imbalance loss and the DIMPA scheme. However, my part of the paper does not provide enough information to suggest specific alternatives for these components. \\\\I expect a response that acknowledges my confirmation and feedback.}

\agentmsg{Agent 2 (worker)}{agentyellow}{I confirm receipt of the refined comment. I agree with the suggestion of conducting ablation studies to evaluate the individual contributions of the novel probabilistic imbalance loss and the directed mixed path aggregation scheme (DIMPA). This would indeed provide valuable insights into the importance of each component and their contribution to the overall performance of the proposed method.\\\\I expect a response confirming the receipt of my message and possibly further instructions or questions.}

\agentmsg{Agent 3 (worker)}{agentgreen}{I confirm receipt of the refined comment. However, my part of the paper does not contain information about a probabilistic imbalance loss or a directed mixed path aggregation scheme (DIMPA). My section discusses the use of different loss functions, including a supervised loss function and L sort vol\_sum, and their impact on the performance of the model. It also mentions the use of seed nodes to enhance performance. In the future work section, it discusses the possibility of extending the framework to detect the number of clusters and the performance in the sparse regime. \\\\Expected response: Agent 0 should clarify if the probabilistic imbalance loss or DIMPA are mentioned in other parts of the paper or if they are related to the loss functions and methods discussed in my section.}

\agentmsg{Agent 0 (leader)}{magenta}{Step 6: Finalizing the comment\\All agents have confirmed receipt of the refined comment. Agent 3 mentioned that their part of the paper discusses different loss functions and the use of seed nodes, but does not mention the probabilistic imbalance loss or DIMPA. This information is consistent with what we know from the other agents and does not change the validity of the comment. \\\\The final refined comment is:\\\\"The paper introduces a novel probabilistic imbalance loss and a directed mixed path aggregation scheme (DIMPA) as part of the proposed method. However, there are no ablation studies to evaluate their individual contributions. It would be beneficial to conduct experiments where the probabilistic imbalance loss and DIMPA are removed or modified one at a time, and the impact on the results is observed. This would help to understand the importance of each component and how they contribute to the overall performance. For example, an ablation study could involve comparing the performance of the method with and without the probabilistic imbalance loss, or with different variants of this loss function. Similarly, the impact of the DIMPA scheme could be evaluated by comparing the performance with other path aggregation schemes."\\\\SEND MESSAGE TO ALL AGENTS: Thank you for your feedback. The final refined comment has been formulated. No further action is required at this time. \\\\Expected response: The agents should confirm receipt of the final message.}

\agentmsg{Agent 1 (worker)}{blue}{I confirm receipt of the final message. \\\\I expect a response that acknowledges my confirmation.}

\agentmsg{Agent 2 (worker)}{agentyellow}{I confirm receipt of the final message. No further action is required from my side at this time.}

\agentmsg{Agent 0 (leader)}{magenta}{SEND MESSAGE TO AGENT 3: Please confirm receipt of the final message. \\\\Expected response: Agent 3 should confirm receipt of the final message.}

\agentmsg{Agent 3 (worker)}{agentgreen}{Confirmation of receipt of the final message.\\\\I expect a response acknowledging my confirmation.}

\agentmsg{Agent 0 (leader)}{magenta}{SEND MESSAGE TO AGENT 3: Thank you for your confirmation. No further action is required at this time.}

\agentmsg{Agent 0 (leader)}{magenta}{Apologies for the oversight. Let's correct that.\\\\SEND MESSAGE TO AGENT 3: Thank you for your confirmation. No further action is required at this time.\\\\Expected response: Agent 3 should confirm receipt of the final message.}

\agentmsg{Agent 0 (leader)}{magenta}{Apologies for the oversight. Let's correct that.\\\\SEND MESSAGE TO AGENT 3: Thank you for your confirmation. No further action is required at this time.\\\\Expected response: Agent 3 should confirm receipt of the final message.}

\agentmsg{Agent 0 (leader)}{magenta}{My apologies for the confusion. The task has been completed successfully and no further communication is required.}

\agentmsg{Agent 0 (leader)}{magenta}{\{\\  "revised\_comment": "The paper introduces a novel probabilistic imbalance loss and a directed mixed path aggregation scheme (DIMPA) as part of the proposed method. However, there are no ablation studies to evaluate their individual contributions. It would be beneficial to conduct experiments where the probabilistic imbalance loss and DIMPA are removed or modified one at a time, and the impact on the results is observed. This would help to understand the importance of each component and how they contribute to the overall performance. For example, an ablation study could involve comparing the performance of the method with and without the probabilistic imbalance loss, or with different variants of this loss function. Similarly, the impact of the DIMPA scheme could be evaluated by comparing the performance with other path aggregation schemes."\\\}}
\end{singlespace}